\documentclass[twoside,11pt]{article}

\usepackage{blindtext}
\usepackage{soul}

\usepackage{jmlr2e}

\usepackage{graphicx}
\usepackage{booktabs}
\usepackage{multirow}
\usepackage{arydshln}
\usepackage{algorithm}     % For algorithm floating environment
\usepackage{algpseudocode} % For pseudocode styling
\usepackage{enumitem}
\usepackage{framed}
\usepackage[export]{adjustbox}
\usepackage[accsupp]{axessibility}  % Improves PDF readability for those with disabilities.
\usepackage{hyperref}
\usepackage{subcaption}
\usepackage{cleveref}

\usepackage{orcidlink}
\algrenewcommand\algorithmicrequire{\textbf{Input:}}
\algrenewcommand\algorithmicensure{\textbf{Output:}}

\newcommand{\method}[0]{\texttt{DECIDER}}
\newcommand{\msp}[0]{\texttt{MSP}}
\newcommand{\ent}[0]{\texttt{Ent}}

\newcommand{\secondbest}[1]{\textcolor{blue}{#1}}
\newcommand\blfootnote[1]{%
  \begingroup
  \renewcommand\thefootnote{}\footnote{#1}%
  \addtocounter{footnote}{-1}%
  \endgroup
}

\ShortHeadings{Leveraging Foundation Model Priors for Improved Model Failure Detection}{DECIDER}
\firstpageno{1}

\begin{document}

\title{DECIDER: Leveraging Foundation Model Priors for Improved Model Failure Detection and Explanation}

% \author{\name Author One \email one@stat.washington.edu \\
%        \addr Department of Statistics\\
%        University of Washington\\
%        Seattle, WA 98195-4322, USA
%        \AND
%        \name Author Two \email two@cs.berkeley.edu \\
%        \addr Division of Computer Science\\
%        University of California\\
%        Berkeley, CA 94720-1776, USA}

\author{\name Rakshith Subramanyam* \email rakshith.subramanyam@axio.ai \\
       \addr Axio.ai \AND
\name Kowshik Thopalli* \email thopalli1@llnl.gov \\
       \addr Lawrence Livermore National Laboratory \AND
    \name Vivek Narayanaswamy* \email narayanaswam1@llnl.gov \\
           \addr Lawrence Livermore National Laboratory 
           \AND
       \name Jayaraman J. Thiagarajan \email jjthiagarajan@gmail.com \\
       \addr Lawrence Livermore National Laboratory 
       }

\editor{}

\maketitle

\begin{abstract}%   <- trailing '%' for backward compatibility of .sty file
Reliably detecting when a deployed machine learning model is likely to fail on a given input is crucial for ensuring safe operation. In this work, we propose DECIDER (Debiasing Classifiers to Identify Errors Reliably), a novel approach that leverages priors from large language models (LLMs) and vision-language models (VLMs) to detect failures in image classification models.
DECIDER utilizes LLMs to specify task-relevant core attributes and constructs a ``debiased'' version of the classifier by aligning its visual features to these core attributes using a VLM, and detects potential failure by measuring disagreement between the original and debiased models. In addition to proactively identifying samples on which the model would fail, DECIDER also provides  human-interpretable explanations for failure through a novel attribute-ablation strategy. Through extensive experiments across diverse benchmarks spanning subpopulation shifts (spurious correlations, class imbalance) and covariate shifts (synthetic corruptions, domain shifts), DECIDER consistently achieves state-of-the-art failure detection performance, significantly outperforming baselines in terms of the overall Matthews correlation coefficient as well as failure and success recall. Our codes can be accessed at~\url{https://github.com/kowshikthopalli/DECIDER/}

  % \keywords{Failure Detection \and Vision-Language Models \and Large-language Models}

\end{abstract}

\blfootnote{\noindent 
* equal contribution}

% Acknowledgements and Disclosure of Funding should go at the end, before appendices and references
\section{Introduction}
A crucial step in ensuring the safety of deployed models is to proactively identify if a model is likely to fail for a given test input. This enables the implementation of appropriate correction mechanisms without impacting the model's operation, or even deferring to human expertise for decision-making. While failures in vision models can be attributed to a variety of factors, the most significant cause is the violation of data distribution assumptions made during training~\citep{jiang2019fantastic}, which is the focus of this work. In general, data comprises both task-relevant \textit{core attributes} and irrelevant \textit{nuisance attributes}, and they are never explicitly annotated. Consequently, models can fail to generalize if (i) the training data contains spurious correlations (to nuisance attributes) that do not appear at test time, (ii) class-conditional distribution of nuisance attributes can arbitrarily change between train and test data (e.g., patient race imbalance in clinical datasets), or (iii) novel attributes emerge only at test time (e.g., style changes). Note that, when the class-conditional distributions of core attributes themselves change between train and test data, it leads to the more challenging scenario of \textit{concept shifts}, and is not considered in this work. Nevertheless, detecting failures across all these scenarios is known to be challenging~\citep{joshi2022all, yang2023change, geirhos2020shortcut}, and hence there has been a surge in research interest~\citep{hendrycks17baseline, guillory2021predicting, gal2016dropout, kirsch2021pitfalls, jain2023distilling}.

We begin by acknowledging that it is not only difficult, but also inefficient, to describe such nuisance attribute discrepancies solely using visual features. In this regard, we explore the utility of large language models (LLMs) and vision-language models (VLMs) in characterizing data attributes through a combination of visual and natural language descriptors. Subsequently, one can leverage these descriptors to design powerful failure detectors that  systematically discern gaps in model generalization. Based on this idea, we develop \method~(\ul{De}biasing \ul{C}lassifiers to \ul{Id}entify \ul{E}rrors \ul{R}eliably), a new approach for failure detection in vision models. At its core, \method~(i) utilizes LLMs (e.g., GPT-3~\citep{brown2020language}) to specify task-relevant core attributes, (ii) uses a VLM (e.g., CLIP~\citep{radford2021learning}) to construct a ``debiased'' version of the task model by aligning its visual features to the core attributes, and (iii) detects failure by measuring disagreement between the original and debiased models for any given test input. 

Additionally, \method~can be used to provide explanations for failure cases. This is done by employing an attribute-ablation strategy that adjusts the relative importance of core attributes such that the prediction probabilities of the debiased matches the original model. Our extensive empirical evaluation shows that our method achieves state-of-the-art performance in detecting failures across various datasets and test scenarios. In summary, our work provides early evidence for the utility of large-scale foundation models as priors for designing novel safety mechanisms.

\section{Related Work}
\noindent\textbf{Failure Detection}. Failure detection in classification involves identifying incorrect predictions made by the model~\citep{hendrycks17baseline, zhu2022rethinking, qu2022improving}. This problem ultimately boils down to identifying an appropriate metric or a \textit{scoring function} that can delineate failed samples from successful ones. Early work involves using simple scores directly derived from the predictions of the model such as Maximum Softmax Probability (MSP)~\citep{hendrycks17baseline}, predictive entropy~\citep{kirsch2021pitfalls} and energy~\citep{liu2020energy} to identify failed samples. More recent work focuses on scores that quantify failure by evaluating the local manifold smoothness~\citep{ng2022predicting} around a given sample and those that are based on agreement of a sample between different components of an ensemble~\citep{jiang2022assessing, trivedi2023closer}. However, such metrics can become unreliable to characterize failure as the model used to derive them can be potentially mis-calibrated and unreliable~\citep{guo2017calibration, minderer2021revisiting}. Failure detection has also been studied under the lens of generalization gap estimation~\citep{guillory2021predicting, narayanaswamygengap} where the goal is to predict the accuracy of the model on an unlabeled target distribution using distributional metrics derived from a number of calibration datasets.

\noindent\textbf{Failure Detection with Vision Language Foundation Models}. Visual-Language Models (VLMs)~\citep{clip, li2022blip} are pre-trained on a large-corpora of image-text captions using a self-supervised objective. VLMs facilitate flexible adaptation to downstream tasks through zero-shot transfer or fine-tuning, demonstrating enhanced performance in zero-shot classification and OOD detection~\citep{wei2023improving, wortsman2022robust, goyal2023finetune, ming2022delving, wang2023clipn, michels2023contrastive, esmaeilpour2022zero}. Recently, VLMs have been used as a lens to understand the failure modes and weaknesses of any pre-trained model. For instance, the authors of~\citep{jain2023distilling} fit a post-hoc failure detector on the latent spaces of the VLM to estimate whether a sample has been correctly identified or not by the pre-trained classifier. The detector is then used to identify the directions of classifier failure modes. However, this approach requires a carefully tailored calibration set to fit the detector which is often unavailable in practice. On the other hand, the authors of~\citep{deng23great} demonstrate that the latent space agreement between the pre-trained model and the VLM is a potential indicator for failure. In contrast, our paper aims to perform failure detection by first designing an improved classifier leveraging the VLM latent space and assessing the agreement between the classifier and its enhanced version while providing explanations for failure.

\begin{figure*}[t]
  \centering
  \includegraphics[width=0.9\textwidth]{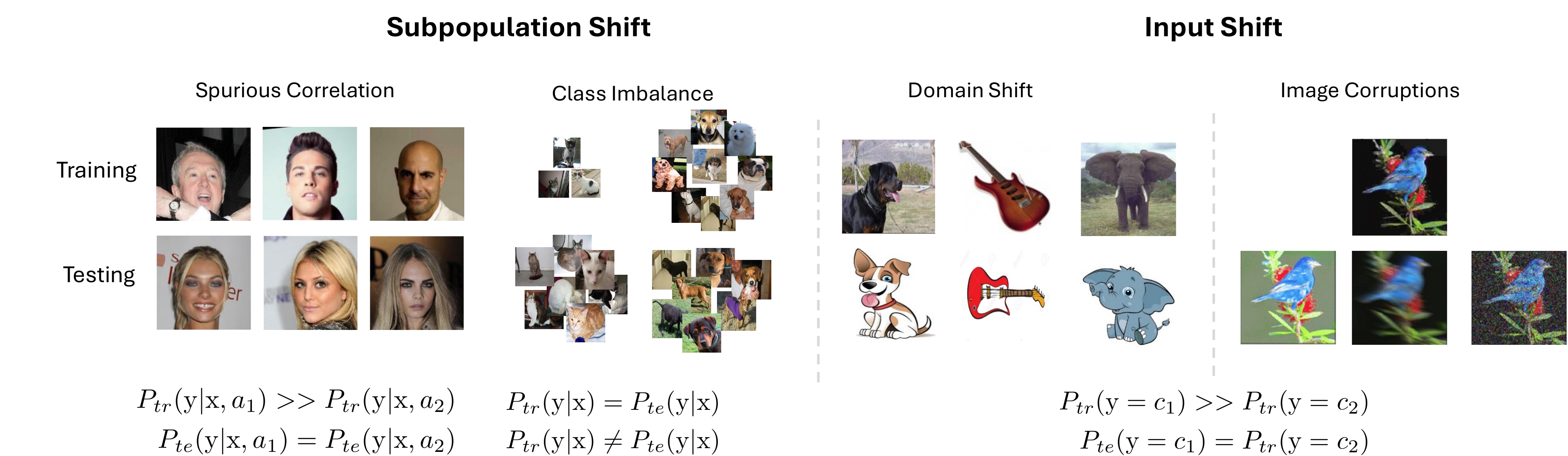} 
  \caption{A visual illustration of the different failure scenarios we consider. These include scenarios when the model relies on spurious correlations present in the data i.e., when an attribute is spuriously correlated with the label (e.g., color of hair and gender). Another cause of failure is when the training data has class imbalance, leading to poorer generalization on images from the under-sampled class. Lastly, another important cause of failures are when the distribution of the test data is different from the training data. This can range from natural image corruptions to covariate shifts.} 
  \label{fig:shifts} 
\end{figure*}

\section{Background}
\label{sec:background}
\noindent\textbf{Preliminaries}. Let $\mathcal{F}$ denote a multi-class classifier with parameters $\theta$, trained on a dataset $ \mathcal{D} = {(\mathrm{x}_i, \mathrm{y}_i)}^{M}_{i=1}$ comprising $M$ samples. Here, $\mathrm{x}_i \in \mathcal{X}$, is a $3$ channel, input RGB image, and $\mathrm{y}_i \in \mathcal{Y}$ is the corresponding label, where $\mathcal{Y}$ is the set of class labels i.e., $\mathcal{Y} = \{1, 2, \ldots, C\}$. Here, $C$ denotes the total number of distinct classes. The classifier $\mathcal{F}$ operates on the input to produce the logits $\mathcal{F}(\mathrm{x})$ corresponding to every class which is followed by a \texttt{softmax} operation to estimate output probabilities $p(\mathrm{y}=c|\mathrm{x})$ where $c$ corresponds to the class index. 

In this paper, we consider the problem of failure detection in classification models, where the source of failure arises due to the following scenarios (Fig. \ref{fig:shifts}) - \underline{(i) Input level shifts} where the training and test images share identical conditional output distributions i.e., $P_{tr}(\mathrm{y}|\mathrm{x}) = P_{te}(\mathrm{y}|\mathrm{x})$ but different input marginals $P_{tr}(\mathrm{x}) \neq P_{te}(\mathrm{x})$. Here, the test data can corresponds to domain variations or image corruptions. \underline{(ii) Sub-population shifts} (a) Spurious correlation where the labels are non-causally associated ~\citep{yang2023change} with certain input characteristics or attributes in the training data over others leading to learning non-generalizable decision rules. For instance, let $a_1$ and $a_2$ correspond to two attributes of an image $\mathrm{x}$ and the training distribution is such that  $P_{tr}(\mathrm{y}|\mathrm{x}, a_1) >> P_{tr}(\mathrm{y}|\mathrm{x}, a_2)$. This model is susceptible to spurious correlations between the inputs and the targets and can fail during test time when  $P_{te}(\mathrm{y}|\mathrm{x}, a_1) = P_{te}(\mathrm{y}|\mathrm{x}, a_2)$, (b) Class imbalance where the number of examples in a given class can be significantly greater than those present in another i.e.,  $P_{tr}(\mathrm{y} = c_1) >> P_{tr}(\mathrm{y} = c_2)$. This does not allow the classifier to optimally capture the image statistics and semantics of class $c_2$ leading to sub-optimal generalization performance.

\noindent\textbf{Failure Detector Design}. Failure detection is a binary classification problem of identifying whether an input sample has been correctly predicted or not by the model. We define our failure detector $\mathcal{G}$ as follows,
\begin{align}
& \mathcal{G}(\mathrm{x}; \theta, \tau) = \begin{cases}
\text{failure}, &\text{if~ $s(\mathrm{x}; \theta) < \tau$},\\
\text{success}, &\text{if~ $s(\mathrm{x}; \theta) \geq \tau$}.
\end{cases}
 \label{eqn:fd_detection}
\end{align}Here, $s(.)$ is a scoring function derived from the classifier $\mathcal{F}$ that assigns higher values for correctly identified samples and vice-versa and $\tau$ is the user-controlled threshold for detection. Following standard practice from the generalization gap literature~\citep{trivedi2023closer, garg2022leveraging}, we identify $\tau$ such that $\sum_{i} \mathbb{I}(s(\mathrm{x}_i; \theta) \geq \tau)$ approximates the true accuracy of the held-out validation dataset. 

\noindent\textbf{Contrastive Language-Image Pre-training (CLIP)}. CLIP~\citep{radford2021learning} is a vision-language model trained on large corpus of image-text pairs with self-supervised learning. It aligns images with natural language descriptions in a shared embedding space, enabling zero-shot learning and fine-tuning for downstream tasks such as image captioning~\citep{subramanyam2023crepe} and visual question answering~\citep{song2022clip, yu2024self, guo2023images, schwenk2022okvqa}. CLIP employs image ($I(.)$) and text ($T(.)$) encoders to generate embeddings ($\mathrm{z}_{I}$ and $\mathrm{z}_{T}$). For zero-shot inference, it computes the cosine similarity ($\cos sim$) between image and text embeddings. This similarity yields class-specific logit scores for zero-shot classification, where the prediction probability $p(\mathrm{y}|\mathrm{x})$ is calculated using \texttt{softmax}.

\begin{figure*}[t]
  \centering
  \includegraphics[width=1\textwidth]{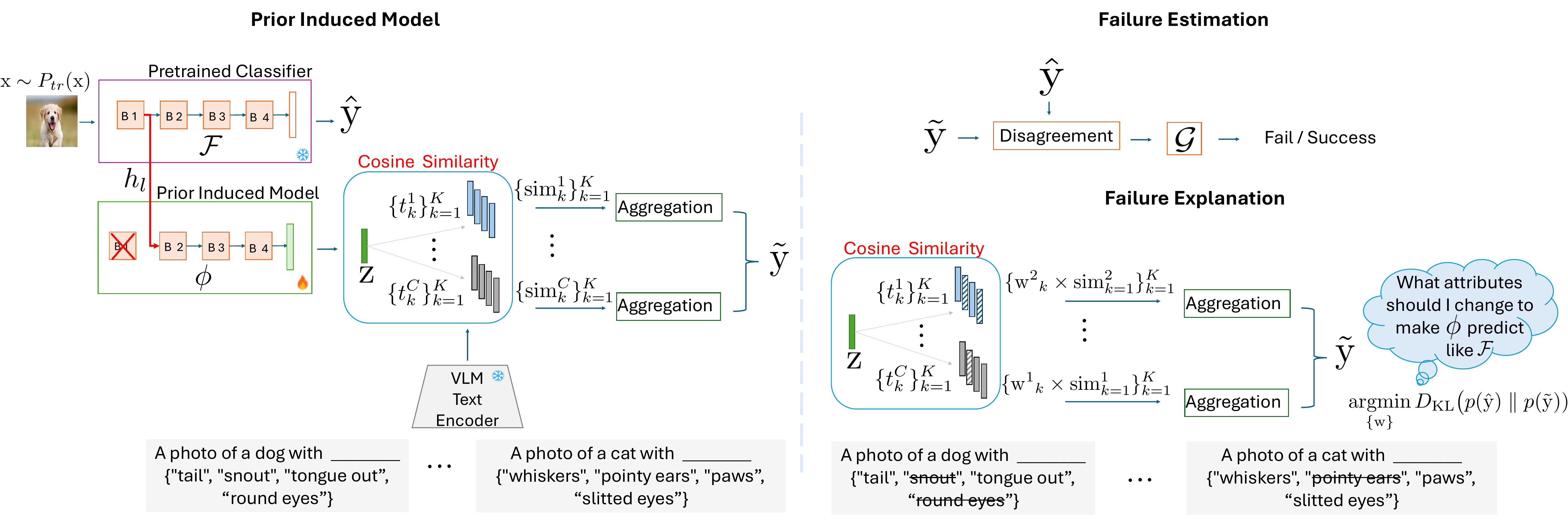} 
  \caption{
\textbf{\method~for failure detection.} \textit{(Left)} \method~trains a Prior Induced Model (PIM) $\phi$, identical to the architecture of the pre-trained classifier $\mathcal{F}$, utilizing priors from a VLM model. 
\textit{(Top Right)} The disagreement between the predictions of $\mathbb{\phi}$ and $\mathcal{F}$ serves as an indicator for failure detection. \textit{(Bottom Right)} By adjusting attribute level weights, \method~offers explanatory insights into failures.
 }
  \label{fig:architecture} 
\end{figure*}

\section{Proposed Approach}
\label{sec:approach}
\subsection{Motivation}

Typically, a classifier $\mathcal{F}$ is trained on a dataset $\mathcal{D}$ to learn the mapping between inputs and target labels. The datasets contain both task-relevant \textit{core attributes} and irrelevant \textit{nuisance attributes}, which are not explicitly annotated. Consequently, the decision rules of the classifier could rely on nuisance attributes leading to poor generalization. For e.g., the model can fail to generalize if the training data contains spurious correlations with nuisance attributes that do not appear during testing. We underscore that this problem of reliance on nuisance attributes arises due to the difficulty in describing them solely using visual features. 

To address this, we go beyond using only visual features and propose to leverage a combination of vision and language descriptors through the use of LLMs and VLMs and design failure detectors that discern the gap in model generalization. In this section we describe our novel strategy for failure detection which involves training a classifier referred to as the Prior Induced Model (PIM) $\mathbb{\phi}$ with the aid of LLMs and VLMs. We believe that the prior knowledge induced by VLMs will help PIM associate task-relevant core attributes. We first describe our paradigm that incorporates foundation models in classifier training. We then develop a prediction disagreement based strategy between PIM and the original classifier to conduct failure detection. Finally, we elucidate the capability of our approach in extracting failure explanations in order to support interpretability.

\subsection{Incorporating Foundation Model Priors}
A key challenge in traditional classification models is the direct mapping of images to coarse labels which encapsulate several attributes. For instance, in distinguishing between a dog and a cat, the label ``dog'' encompasses attributes like ``wagging tail'' and ``snout'', while ``cat'' includes ``whiskers'' and ``pointy ears''. Without explicit access to such detailed attribute information and due to potential biases in the training data, models are susceptible to rely on overly simplistic decision rules. In contrast, VLMs such as CLIP offer capabilities to encode both image and textual attribute descriptions into a unified latent space that is enriched to support meaningful image-text attribute associations. 

To improve the effectiveness of classification model training, we hypothesize that aligning the model's visual features with the textual descriptions of core attributes related to the class of interest in the VLM latent space can enhance training. This alignment is expected to equip the classifier with the ability to develop decision-making rules that are both more reliable and generalizable, while also reducing the influence of existing biases.

To achieve this, we introduce the PIM model $\mathbb{\phi}$, which is guided by the LLM and VLM based priors (see Fig. \ref{fig:architecture} left). The architecture of PIM closely resembles that of its counterpart $\mathcal{F}$, with the notable distinction being that its final layer projects onto the VLM latent space. This projection supports the alignment with the textual descriptions of class-level attributes, thereby harnessing the linguistic capabilities of foundational models. PIM is specifically engineered to accept early-stage features from $\mathcal{F}$, denoted as $h_l$, which are then processed through PIM's analogous layers to produce the image encoding $\mathrm{z}$ within the VLM latent space. For instance, when both $\mathcal{F}$ and $\mathbb{\phi}$ are based on the ResNet architecture ~\citep{he2016deep}, the output from block 1 of $\mathcal{F}$ serves as the input for block 2 in $\mathbb{\phi}$.

It must be noted that the success of our approach relies upon the quality of the fine-grained text attributes extracted for every class. While there exists strategies~\citep{merullo2022linearly} that are capable of extracting image-level textual descriptions, they usually involve the text decoders in the loop which can be computationally expensive. Therefore, we resort to using Large Language Models (LLMs) to compute task-specific attribute descriptions offline. 

\subsection{Generating Task-specific Core-attribute Descriptions}LLMs~\citep{touvron2023llama, brown2020language} have demonstrated their utility across a range of language tasks\citep{radford2019language, wei2022finetuned, nakano2021webgpt, pratt2023does} and are particularly adept at contextual understanding, and generating coherent text even with descriptive prompting. To extract the class-specific attribute descriptions,  we query GPT-3~\citep{brown2020language} with the prompts ``List visually descriptive attributes of <\texttt{CLASS}>.'' This allows us to gather a set of $K$ attributes $\mathcal{A}^c = \{a^c_k \}_{k=1}^K $ for every class $c$.

\subsection{Training PIM}
\noindent\textbf{(i) Computing Cosine Similarities}. We first compute the cosine similarity scores between the image embedding $\mathrm{z}$ produced by PIM for a given image and the text embeddings associated with attribute $k$ from each class $c$. It is given by, 
\begin{equation}
\Omega_{\mathcal{A}^c} = \{ \omega^c_k \}_{k=1}^K \text{ where } \omega^c_k = \cos sim(\mathrm{z}, e^c_k)
\end{equation}
Here, the text embeddings $E_{\mathcal{A}^c} = \{ e^c_k \}_{k=1}^K$ for each attribute of every class are obtained using the CLIP text encoder. 

\noindent\textbf{(ii) Attribute Similarity Aggregation}.
Subsequently, we aggregate these attribute similarity scores, $\Omega_{\mathcal{A}^c}$, for each class $c$ to obtain coarse prediction logits corresponding to the class label $\mathrm{y} \in \mathcal{Y}$. We investigate two aggregation strategies namely - (i) Class-level mean and (ii) Class-level max to consolidate these scores into final class predictions which are eventually normalized using $\texttt{softmax}$. These strategies enable a more refined and attribute-aware determination of classification outcomes.

\noindent\textbf{(iii) Optimization Objective}.
The optimization is primarily guided by the cross-entropy loss which evaluates the discrepancy between the predicted probabilities from PIM and the ground truth label. In addition, we include consistency driven augmentations namely CutMix~\citep{yun2019cutmix} and AugMix~\citep{hendrycks2020augmix} to improve its robustness. Additionally, we upweight the losses corresponding to the instances where (i) the biased classifier $\mathcal{F}$ predicts accurately, but $\mathbb{\phi}$ does not and (ii) the biased classifier $\mathcal{F}$ does not predict accurately, as well as $\mathbb{\phi}$ does not, within a training batch. 

\subsection{\method: Failure Estimation Using PIM}
To assess the failure of the biased classifier $\mathcal{F}$, we compute the disagreement between PIM and $\mathcal{F}$ based on the discrepancy between their predictions. This disagreement score is calculated as the cross-entropy between the sample-level probability distributions between the two models with PIM being the reference distribution given by \(
s(\mathrm{x}) = -\sum_{c=1}^C p(\mathrm{y} = c|\mathrm{x}). \log(q(\mathrm{y} = c|\mathrm{x}))
\)
where $p(.)$ and $q(.)$ represent the predicted probabilities of $\mathcal{F}$ and PIM, respectively. 

\subsection{Extracting Explanations for Failure}
Our failure explanation protocol is designed to elucidate the underlying reasons behind the discrepancies between predictions of $\mathcal{F}$ and $\phi$. The primary objective is to identify the optimal subset of attributes necessary for aligning the PIM's prediction probabilities with those of the task model. To achieve this, we implement an attribute ablation strategy where we iteratively adjust a group of weights corresponding to each attribute across all classes. Our iterative process begins by initially assigning uniform weights to every attribute for each class within a batch. These weights are then optimized by minimizing the Kullback-Leibler (KL) divergence between the probability distributions predicted by $\mathcal{F}$ and those adjusted by PIM, accounting for the influence of the weighted attributes.  As the algorithm converges, the weights will highlight those attributes that have significant impact on the predictions of $\mathcal{F}$, providing insights into the features considered by $\mathcal{F}$ when making decisions. Fig. \ref{fig:architecture} right illustrates our failure explanation mechanism.

\section{Empirical Evaluation}
We conduct comprehensive evaluations of \method~using various classification benchmarks and assess performance under various failure scenarios with different architectures. We employ OpenAI's CLIP ViT-B-32 model in all experiments~\citep{clip}. 

\subsection{Experimental Setup}
\noindent\textbf{Datasets}. Our experiments are centered around datasets reflecting four common sources of model failure: 
\renewcommand{\labelitemi}{\textbullet}
\begin{itemize}
    \item \textbf{Input-Level Shifts}: CIFAR100-C~\citep{cifar100-C}, comprising 19 types of corruptions at five severity levels over the CIFAR100 test images across 100 categories.
    \item \textbf{Spurious Correlations}: (1) Waterbirds~\citep{yang2023change} involves classifying images as `water bird' or `land bird'. The training data offers biases tied to the background (water/land). (2) CelebA~\citep{liu2015faceattributes, yang2023change} involves classifying if individuals have blond hair or not, with labels spuriously correlated with gender.
    \item \textbf{Class Imbalance}: We modify the Kaggle Cats vs Dogs dataset~\citep{dogs-vs-cats}, adjusting the distribution to create a training imbalance with 5,989 cat and 19,966 dog images for training, while maintaining balanced test data.
    \item \textbf{Distribution Shifts}:  (1) PACS~\citep{PACS} includes images from four domains (Photo, Art-painting, Cartoon, Sketch), to be classified into seven categories. As two large-scale benchmarks, we consider (2) DomainNet~\citep{domainnet} which contains images from $345$ categories from $6$ domains (Real, Painting, Infograph, Quickdraw, Cartoon and Sketch) and (3) ImageNet-Sketch~\citep{imagenet-sketch} benchmark which contains sketch images from $1000$ ImageNet~\citep{imagenet} classes. 
\end{itemize}

\noindent\textbf{Model Architectures}. We consider the ResNet-50 architecture for CelebA dataset and for all other datasets, we employ ResNet-18 trained on their respective datasets as the original classifier $\mathcal{F}$. In the supplementary, we study the performance of \method~on more architectures and we provide additional training details. 

\subsection{Baselines}
We consider different baselines that use sample-level scores $s$ for failure estimation:- (i) Maximum Softmax Probability ($\texttt{MSP}$)~\citep{hendrycks17baseline} which is given by $s(\mathrm{x}) = \underset{j}{\text{max}} ~p(\mathrm{y}=j|\mathrm{x})$, (ii) Predictive Entropy ($\texttt{Ent}$) is essentially the entropy among the predictions of a sample and is given by $s(\mathrm{x}) = -\sum_{j=1}^{K} p(\mathrm{y}=j|\mathrm{x}).\text{log}(p(\mathrm{y}=j|\mathrm{x}))$, (iii) ~$\texttt{Energy}$~\citep{liu2020energy} score is defined by $s(\mathrm{x}) = -T. \text{log}\sum_{j=1}^{K} \text{exp}^{\mathcal{F}_\theta(\mathrm{x}_j)}$. Following standard practice, we consider $T=1$ in all our experiments. (iv) Generalized Model Disagreement (\texttt{GDE})~\citep{jiang2022assessing, chen2021detecting} - Let $\mathcal{F}_{\theta_1}, \mathcal{F}_{\theta_2} \dots \mathcal{F}_{\theta_r}$ denote $r$ models trained with different random seeds. Let $\mathcal{F}_{\theta_1}$ denote the base classifier. Then the score is computed as $s(\mathrm{x}) = \frac{1}{r}\sum_{i=1}^{r}\frac{1}{r-1}\sum_{j \neq i}^{r} \mathbb{I}(\mathcal{F}_{\theta_i} \neq \mathcal{F}_{\theta_j})$. We set $r$ to $5$.

It must be noted that we utilize negative versions of entropy and energy to reflect the fact the samples that are correctly predicted are associated with higher scores. 

\begin{figure*}[t]
  \centering
  \includegraphics[width=1\textwidth]{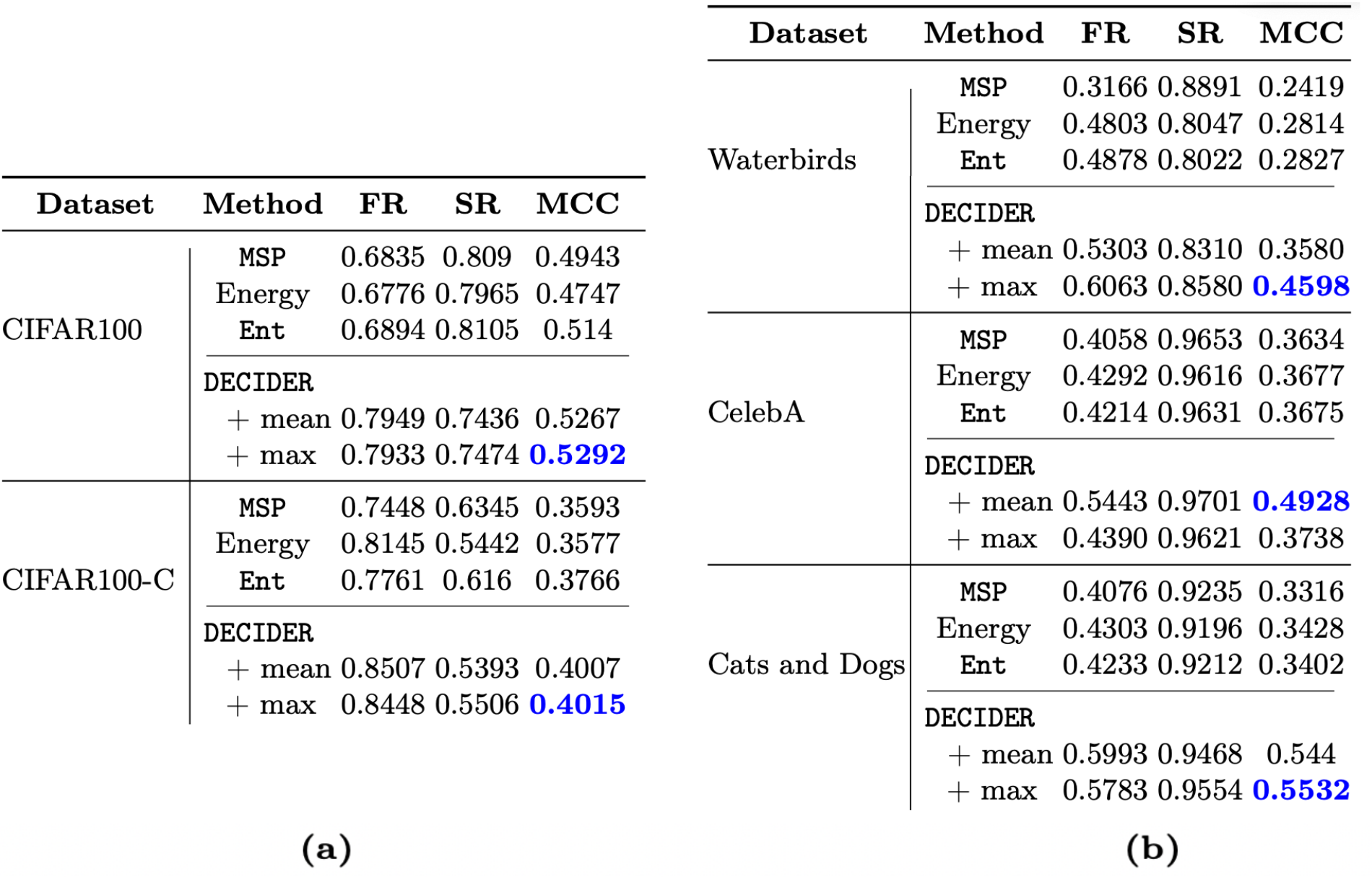} 
  \caption{
  Results on failure detection across different benchmarks - (a) CIFAR100, and image corruptions on CIFAR-100-C, and (b) subpopulation shifts from spurious correlations on Waterbirds, CelebA datasets, and class imbalance on Cats vs Dogs. \method~consistently outperforms baselines in terms of the overall Matthew's Correlation Coefficient (MCC) as well as achieving higher failure and success recalls.} 
  \label{tab:cifar_waterbirds} 
\end{figure*}

\subsection{Metrics}
We consider the following metrics to evaluate failure detection performance: \underline{(i) Failure Recall (FR)} which corresponds to the fraction of samples that have been correctly identified as failure, \underline{(ii) Success Recall (SR)} corresponds to the fraction of samples that have been correctly predicted as successful. The trade-off between the two metrics is indicative of how aggressive or conservative the failure detector is. \underline{(iii) Matthew's Correlation Coefficient (MCC)} holistically assesses the quality of the binary classification task of failure detection and provides a balanced measure when the class sizes are different. It takes into account both true and false positives and negatives respectively while assessing performance.

\subsection{Findings}
\noindent\textbf{Input Shifts}. Fig.~\ref{tab:cifar_waterbirds}(a) showcases the results on the CIFAR100 and CIFAR100-C datasets. On the clean CIFAR100, \method~outperforms the baselines with a superior MCC of $0.5292$ for the max variant(versus $0.514$ for the best baseline), attributed to higher failure recall ($0.7933$) and success recall ($0.7474$). On the more challenging CIFAR100-C (severity level 4), \method~further highlights its efficacy by achieving an MCC of $0.4015$ with max aggregation, exceeding the top baseline (entropy) which has an MCC of $0.3766$. This is due to a balanced trade-off between failure recall ($0.8448$) and success recall ($0.5506$), distinguishing \method~from other baselines that fail to maintain such balance. These findings clearly demonstrate \method~as robust in detecting classifier failures amid input-level shifts, surpassing other baselines in performance metrics.\\

\noindent\textbf{Subpopulation Shifts}. Our comprehensive evaluation addresses datasets affected by various subpopulation shifts. The summarized results in Fig.~\ref{tab:cifar_waterbirds}(b) underline the effectiveness of \method~in navigating these challenges:

\noindent\underline{Waterbirds}: \method~achieves a high failure recall of $0.6063$, outperforming the best baseline (entropy) which has a recall of $0.4878$. Importantly, 
\method~maintains a high success recall ($0.858$) with minimal compromise compared to MSP ($0.8891$). The outcome is a leading MCC of $0.4598$, attesting to \method's balanced detection ability in environments with misleading background cues.

\noindent\underline{CelebA}: With mean aggregation, \method~delivers the highest MCC of $0.4928$, combining a failure recall of $0.5443$ with a success recall of $0.9701$, showcasing its strength in addressing gender and hair color spurious correlations.

\begin{figure}[t]
    \centering
    \begin{subfigure}{0.45\linewidth}
        \centering
        \includegraphics[width=\linewidth]{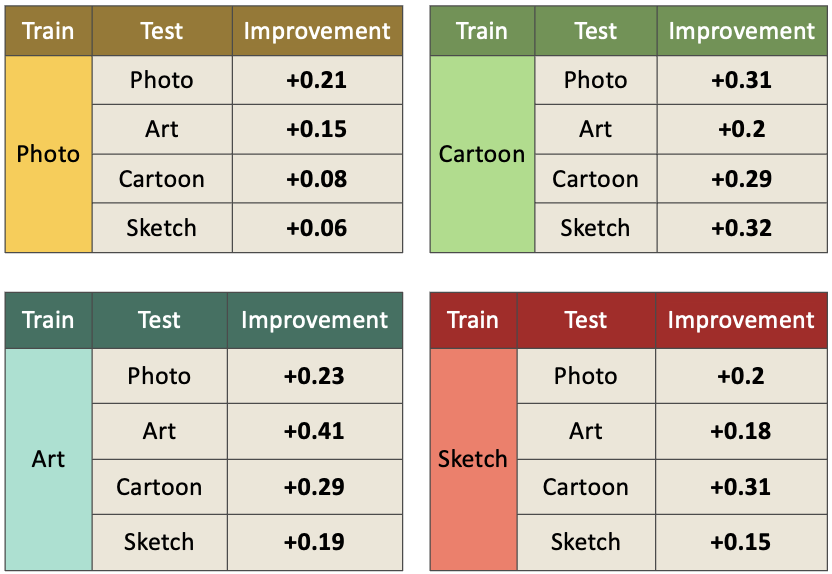}
        \caption{}
        \label{fig:pacs}
    \end{subfigure}
    \hfill
    \begin{subfigure}{0.5\linewidth}
        \centering
        \includegraphics[width=\linewidth]{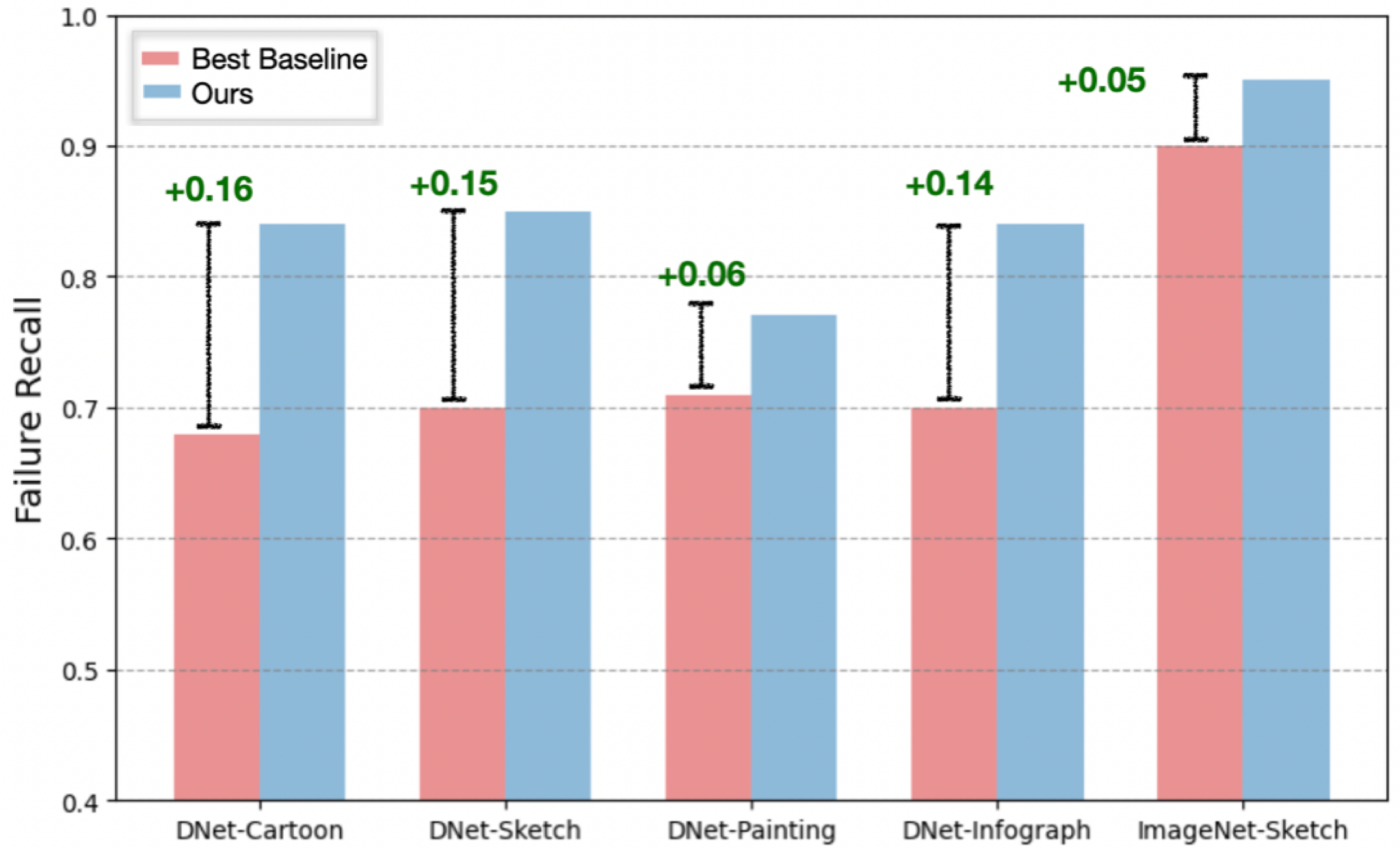}
        \caption{}
        \label{fig:domainnet}
    \end{subfigure}
\caption{\textbf{\method~produces the best performance on covariate shifts.}. (left) Comparison of \method~against the best baseline in terms of the difference in MCC on the PACS dataset involving covariate shifts across 4 different visual domains. (Right) Improvement in failure recall performance of the best performing baseline and \method~on large-scale covariate shift benchmarks- DomainNet (DNet) and ImageNet-Sketch. The classifiers and PIMs are trained on DomainNet Real and Imagenet train sets respectively and evaluated on the different distribution shift datasets.}
    \label{fig:pacs-domainnet}
\end{figure}
\noindent\underline{Cats vs Dogs}: Exhibiting strong performance in class imbalance, \method~(max aggregation) achieves an MCC of $0.5532$, significantly surpassing the top baseline (energy) with an MCC of $0.3428$, underlining its efficacy in balanced success and failure recall. \method~not only demonstrates high failure detection capability but also ensures high success recall rates above $0.94$, highlighting its proficiency in class-imbalanced settings.\\

\noindent\textbf{Covariate Shifts}.
In this section, we evaluate the performance of \method~in the challenging setting of identifying failure due to covariate shifts. We first consider the PACS dataset which contains 4 different domains. We train PIM and derive individual thresholds for each of the four domains and evaluate its performance across all domains. While we present detailed results for baselines and metrics in the supplementary, in Fig.~\ref{fig:pacs-domainnet}(a), we report the gain in MCC scores between the best performing baseline and \method. It can be seen that \method~outperforms the baselines by a large margin across all the domains. To further validate the effectiveness of \method, we conducted experiments on large-scale covariate shift benchmarks, including DomainNet and ImageNet. In the DomainNet case, we trained the classifier and PIM on images from the real domain and evaluated their performance on four different target domains: Cartoon, Sketch, Painting, and Infograph. For ImageNet, we trained on the ImageNet training dataset and assessed the performance on the challenging ImageNet-Sketch benchmark. Fig.~\ref{fig:pacs-domainnet}(b) presents the failure recall performance of the best-performing baseline and \method, clearly demonstrating the superiority of our approach even when applied to large-scale datasets. 

In summary, these results highlight the importance of leveraging language priors together with priors from the VLM to construct debiased models that reliably help detect failures across different scenarios. 

\section{Failure Explanation}
\begin{figure*}[t]
    \centering
    \includegraphics[width=1\linewidth]{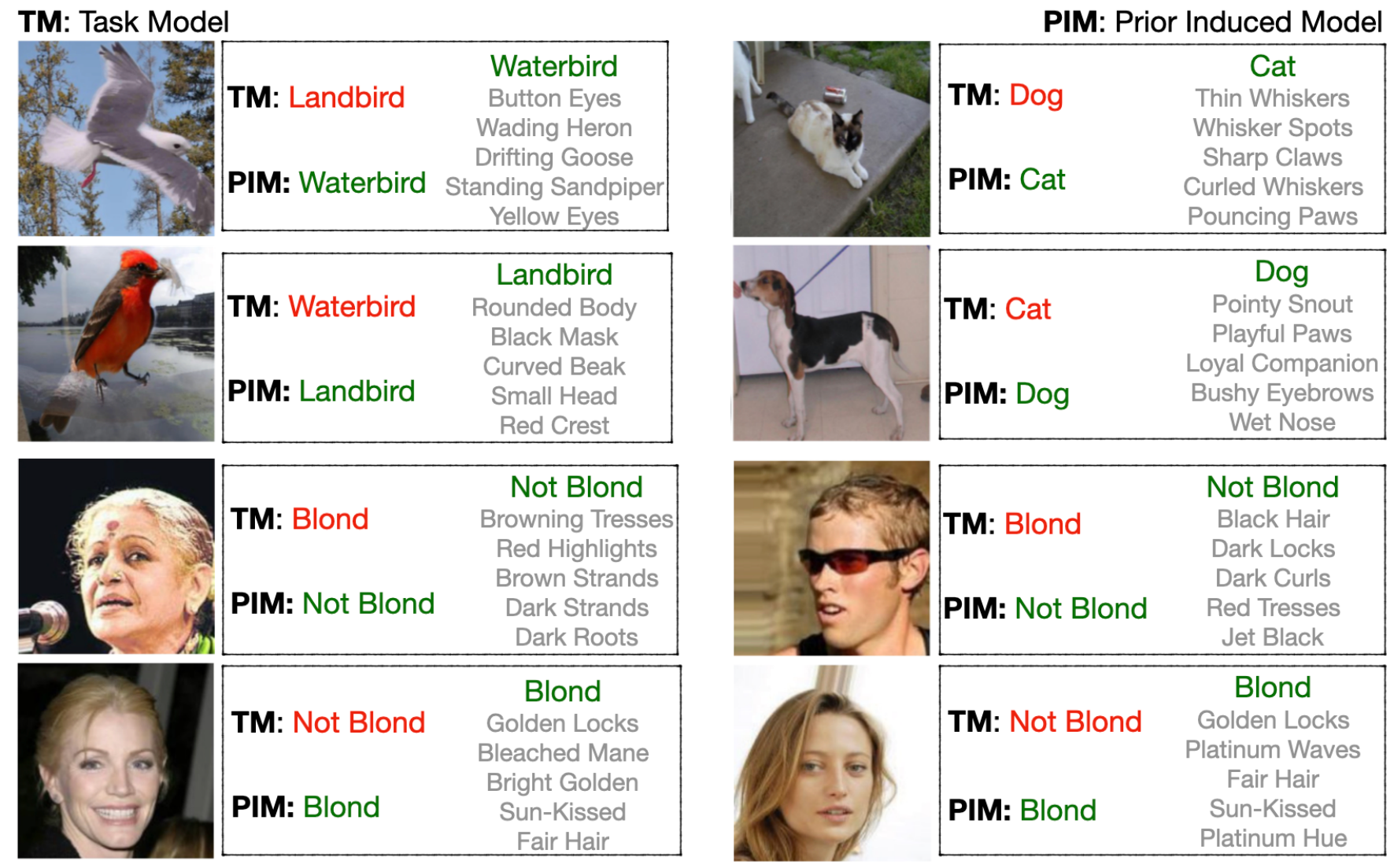}
    \caption{\textbf{Failure Explanations}. We explain the failures of the biased classifier $\mathcal{F}$, by manipulating the influence of individual attributes in PIM, such that the prediction probabilities of PIM match that of $\mathcal{F}$. The knowledge of the attributes whose influence was needed to be reduced provides an indication that $\mathcal{F}$ has not focused on those attributes to make its decisions. We show qualitative examples on Water birds in top left, Cats vs dogs in top right and from CelebA dataset in bottom.}
    \label{fig:explanation}
\end{figure*}

Having empirically demonstrated the superior failure detection capabilities of \method, we now turn our attention to the task of explaining the reasons behind failures. To that end, we consider the max variant of \method~and adjust the influence of individual attributes to ensure that the prediction probabilities generated by \method~closely mirror those of the original model as explained in Section~\ref{sec:approach}. This manipulation offers evidence of what attributes the task model uses. For e.g., on the top left of Fig. \ref{fig:explanation}, the task is to correctly identify the hair color. Here, the classifier $\mathcal{F} $ incorrectly classifies the image, while PIM accurately identifies the same. We observe that our optimization process reduces the influence of core attributes such as "Browning Tresses" and "Red Highlights" on PIM's predictions. This manipulation serves as evidence that the biased classifier $\mathcal{F}$ may not have considered these crucial attributes in its decision-making process. Similarly, in the example shown in Fig. \ref{fig:explanation}, $\mathcal{F}$ misclassifies a Cat as a Dog (top right) and the proposed optimization shows that the classifier has not focused enough on the important core attributes such as ``Thin Whiskers'' thus making the erroneous classification.

\section{Analyses}
\noindent\textbf{Biases or insufficiency of GPT-3 attributes}. The success of \method~relies on the quality of the attributes generated by the LLM. To study the impact on failure detection on the quality of text attributes, we consider two practical scenarios: (i) \ul{GPT-3 generates irrelevant attributes}: In this case, the PIM model has the risk of learning noisy decision rules that the even the classifier might not have; (ii) \ul{GPT provides insufficient attributes}: With only partial attributes,  PIM's predictive performance can be limited. To comprehensively evaluate the impact of both scenarios, we employ the following protocol on the Waterbirds dataset. For scenario (i), we add 5 randomly sampled core attributes from the other class to the attribute set of each class. For case (ii), we remove 5 randomly selected attributes from the attribute set of each class. We train PIM under both these scenarios. As the results in Table~\ref{tab:att_quality} show, although there is a noticeable drop in performance due to the severe attribute corruptions, \method~still outperforms the best baseline (\ent) method.  This demonstrates the robustness of \method~to imperfect attribute sets.
\begin{table}[htbp]
\centering
\caption{\textbf{  Impact of attribute quality} -- (i) \textit{irrelevant}: add 5 nuisance attributes; (ii) \textit{insufficient}: remove 5 core attributes. Although there is a drop in performance under attribute corruptions, \method~still outperforms existing baselines.
}
\renewcommand{\arraystretch}{1.5}
\resizebox{\columnwidth}{!}{
\begin{tabular}{c|c|c|c|c}
\hline
\textbf{Metric} & \textbf{Baseline (\ent)} & \textbf{\method }& \textbf{\method~(irrelevant)} & \textbf{\method~(insufficient)} \\ \hline
Failure Recall  & 0.48              & {\color[HTML]{3531FF}\textbf{0.60} }               & 0.54                         & 0.49                           \\ \hline
Success Recall  & 0.80              & {\color[HTML]{3531FF}\textbf{0.85}}             & 0.81                         & 0.83                           \\ \hline
MCC             & 0.28              & {\color[HTML]{3531FF}\textbf{0.45}}                & 0.34                         & 0.33                           \\ \hline
\end{tabular}
}
\label{tab:att_quality}
\end{table}

\begin{figure*}[h]
  \centering
  \includegraphics[width=1\textwidth]{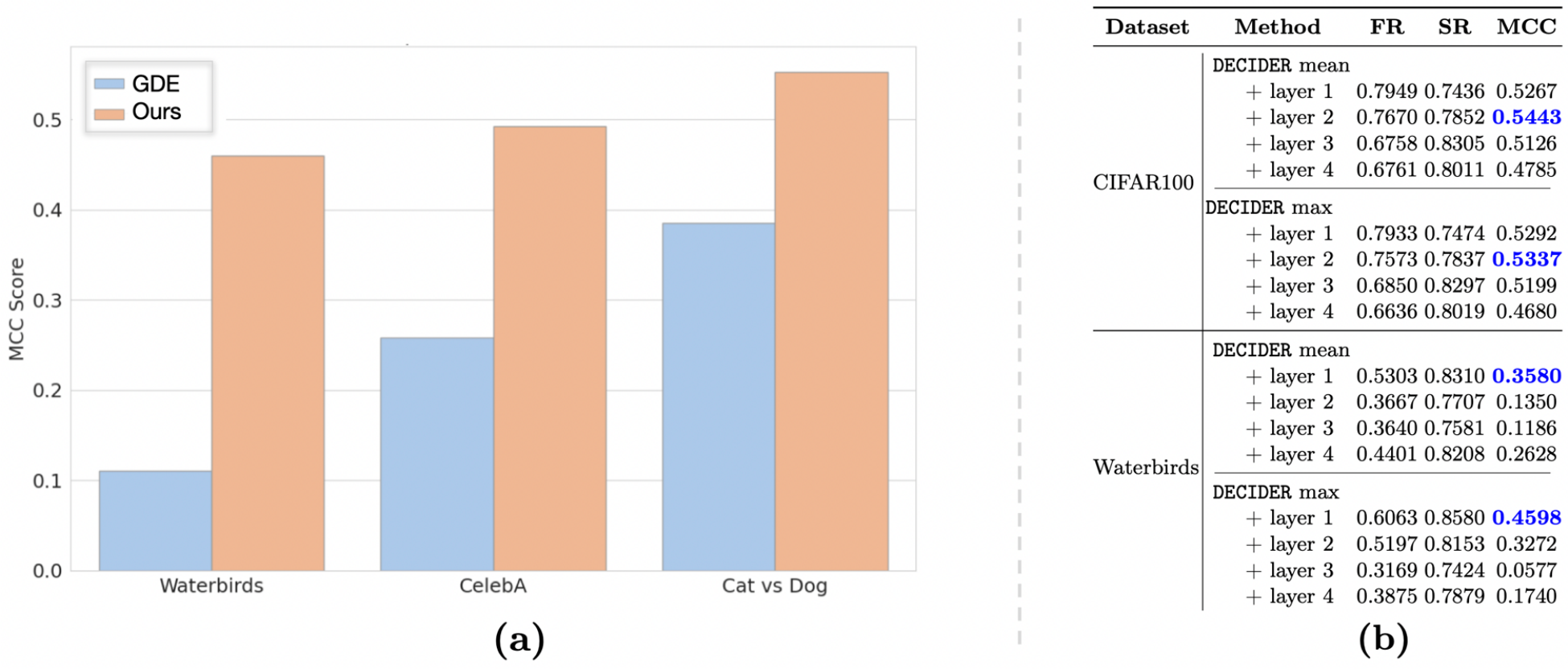} 
  \caption{ 
  (a) Comparison of \method~against the failure detection performance obtained through disagreement between predictions from an ensemble of multiple instances of $\mathcal{F}$ on  Waterbirds, CelebA and Cats vs Dogs datasets respectively. (b) Ablation study analyzing the impact of using features from different layers of the base model $\mathcal{F}$ as input to the Prior Induced Model (PIM) $\phi$ on CIFAR-100 and Waterbirds datasets.} 
  \label{fig:gde} 
\end{figure*}

\noindent{\textbf{Impact of Layer Selection of $\mathcal{F}$ on $\phi$}}. In this study, we explore how the performance of the PIM model $\phi$ is influenced by the specific layer in $\mathcal{F}$ from which we extract features. This experiment uses the ResNet-18 architecture, with models trained on the CIFAR100 and Waterbirds datasets. From the results presented in the table in Fig. \ref{fig:gde}, using features from the early layers (layer 1 and layer 2) of ResNet-18 yields the highest MCC (Matthews Correlation Coefficient) scores. In contrast, leveraging features from the later layers leads to a noticeable decline in performance. This observation suggests that the initial layers of the network are less prone to carrying biases than the later ones, supporting the findings from previous research~\citep{lee2022surgical}.

\noindent{\textbf{Model Ensembles for Disagreement Analysis}}. 
It has been shown that the prediction disagreement between different constituent members of a model ensemble can serve as an indicator of failure~\citep{jiang2022assessing, trivedi2023closer}. In this experiment, we compare the failure estimation performance obtained through the disagreement between PIM and $\mathcal{F}$ to the performance obtained by the disagreement between an ensemble (GDE). To that end, we trained five different classifiers with different initial seeds on three different datasets: Waterbirds, CelebA, and Cat vs Dogs. 
Figure~\ref{fig:gde}, evidences the superiority of the proposed approaches compared to GDE. 

\noindent{\textbf{Impact of PIM accuracy on failure detection}}. Since we attempt to train a debiased classifier, in this section, we study the impact of its accuracy on failure detection. Table~\ref{tab:pim_acc} in the appendix reveals that, despite the occasional slight decrease in the predictive performance of the debiased model PIM, the core-nuisance attribute disambiguation, which is crucial for failure detection, is not compromised. Consequently, \method~consistently achieves superior failure recall compared to the baselines.

\noindent{\textbf{Replacing PIM with CLIP classifiers}}. Given that we propose leveraging the priors from CLIP to obtain a debiased version of the classifier, it is natural to consider utilizing CLIP's zero-shot classifier directly as PIM. Table~\ref{tab:CLIP_PIM} in appendix demonstrates that such an approach yields poor failure detection performance when CLIP's zero-shot classifier is employed as PIM. This is because the visual features and their correlations to the core attributes of CLIP can differ significantly from the task model, thus rendering the model disagreement based failure detection highly ineffective.

\section{Conclusion}
In this work, we introduced \method, a novel approach that leverages LLMs and vision-language foundation models to detect failures in pre-trained image classification models.  Our key insight was to train an improved version of the pre-trained classifier, PIM, that learns robust associations between visual features and class-level attributes by projecting into the shared embedding space of a VLMs such as CLIP. By analyzing the disagreement between PIM's predictions and the original biased model, \method~can reliably identify potential failures while offering human-interpretable explanations. Extensive experiments across multiple benchmarks evidences the consistent superiority of \method~over baselines, achieving substantially higher overall scores and better trade-offs between failure and success recalls. Our work highlights the promise of integrating vision-language priors into model failure analysis pipelines to facilitate more reliable and trustworthy deployment of vision models in safety-critical applications. Extending \method~to other vision-language models and exploring its application to other failure modes such as adversarial attacks constitute our future work.

\acks{This work was performed under the auspices of the U.S. Department of Energy by the Lawrence Livermore National Laboratory under Contract No. DE-AC52-07NA27344, Lawrence Livermore National Security, LLC. Supported by LDRD project 24-FS-002. LLNL-CONF-862086.}

\bibliography{main}

\begin{thebibliography}{54}
\providecommand{\natexlab}[1]{#1}
\providecommand{\url}[1]{\texttt{#1}}
\expandafter\ifx\csname urlstyle\endcsname\relax
  \providecommand{\doi}[1]{doi: #1}\else
  \providecommand{\doi}{doi: \begingroup \urlstyle{rm}\Url}\fi

\bibitem[Brown et~al.(2020)Brown, Mann, Ryder, Subbiah, Kaplan, Dhariwal, Neelakantan, Shyam, Sastry, Askell, et~al.]{brown2020language}
Tom Brown, Benjamin Mann, Nick Ryder, Melanie Subbiah, Jared~D Kaplan, Prafulla Dhariwal, Arvind Neelakantan, Pranav Shyam, Girish Sastry, Amanda Askell, et~al.
\newblock Language models are few-shot learners.
\newblock \emph{Advances in neural information processing systems}, 33:\penalty0 1877--1901, 2020.

\bibitem[Chen et~al.(2021)Chen, Liu, Avci, Wu, Liang, and Jha]{chen2021detecting}
Jiefeng Chen, Frederick Liu, Besim Avci, Xi~Wu, Yingyu Liang, and Somesh Jha.
\newblock Detecting errors and estimating accuracy on unlabeled data with self-training ensembles.
\newblock \emph{Advances in Neural Information Processing Systems}, 34:\penalty0 14980--14992, 2021.

\bibitem[Cukierski(2013)]{dogs-vs-cats}
Will Cukierski.
\newblock Dogs vs. cats, 2013.
\newblock URL \url{https://kaggle.com/competitions/dogs-vs-cats}.

\bibitem[Deng et~al.(2023)Deng, Xiong, and Hooi]{deng23great}
Ailin Deng, Miao Xiong, and Bryan Hooi.
\newblock Great models think alike: Improving model reliability via inter-model latent agreement.
\newblock In \emph{Proceedings of the 40th International Conference on Machine Learning}, volume 202 of \emph{Proceedings of Machine Learning Research}, pages 7675--7693. PMLR, 23--29 Jul 2023.

\bibitem[Esmaeilpour et~al.(2022)Esmaeilpour, Liu, Robertson, and Shu]{esmaeilpour2022zero}
Sepideh Esmaeilpour, Bing Liu, Eric Robertson, and Lei Shu.
\newblock Zero-shot out-of-distribution detection based on the pre-trained model clip.
\newblock In \emph{Proceedings of the AAAI conference on artificial intelligence}, volume~36, pages 6568--6576, 2022.

\bibitem[Gal and Ghahramani(2016)]{gal2016dropout}
Yarin Gal and Zoubin Ghahramani.
\newblock Dropout as a bayesian approximation: Representing model uncertainty in deep learning.
\newblock In \emph{international conference on machine learning}, pages 1050--1059. PMLR, 2016.

\bibitem[Garg et~al.(2022)Garg, Balakrishnan, Lipton, Neyshabur, and Sedghi]{garg2022leveraging}
Saurabh Garg, Sivaraman Balakrishnan, Zachary~Chase Lipton, Behnam Neyshabur, and Hanie Sedghi.
\newblock Leveraging unlabeled data to predict out-of-distribution performance.
\newblock In \emph{International Conference on Learning Representations}, 2022.
\newblock URL \url{https://openreview.net/forum?id=o_HsiMPYh_x}.

\bibitem[Geirhos et~al.(2020)Geirhos, Jacobsen, Michaelis, Zemel, Brendel, Bethge, and Wichmann]{geirhos2020shortcut}
Robert Geirhos, J{\"o}rn-Henrik Jacobsen, Claudio Michaelis, Richard Zemel, Wieland Brendel, Matthias Bethge, and Felix~A Wichmann.
\newblock Shortcut learning in deep neural networks.
\newblock \emph{Nature Machine Intelligence}, 2\penalty0 (11):\penalty0 665--673, 2020.

\bibitem[Goyal et~al.(2023)Goyal, Kumar, Garg, Kolter, and Raghunathan]{goyal2023finetune}
Sachin Goyal, Ananya Kumar, Sankalp Garg, Zico Kolter, and Aditi Raghunathan.
\newblock Finetune like you pretrain: Improved finetuning of zero-shot vision models.
\newblock In \emph{Proceedings of the IEEE/CVF Conference on Computer Vision and Pattern Recognition}, pages 19338--19347, 2023.

\bibitem[Guillory et~al.(2021)Guillory, Shankar, Ebrahimi, Darrell, and Schmidt]{guillory2021predicting}
Devin Guillory, Vaishaal Shankar, Sayna Ebrahimi, Trevor Darrell, and Ludwig Schmidt.
\newblock Predicting with confidence on unseen distributions.
\newblock In \emph{Proceedings of the IEEE/CVF international conference on computer vision}, pages 1134--1144, 2021.

\bibitem[Guo et~al.(2017)Guo, Pleiss, Sun, and Weinberger]{guo2017calibration}
Chuan Guo, Geoff Pleiss, Yu~Sun, and Kilian~Q Weinberger.
\newblock On calibration of modern neural networks.
\newblock In \emph{International conference on machine learning}, pages 1321--1330. PMLR, 2017.

\bibitem[Guo et~al.(2023)Guo, Li, Li, Tiong, Li, Tao, and Hoi]{guo2023images}
Jiaxian Guo, Junnan Li, Dongxu Li, Anthony Meng~Huat Tiong, Boyang Li, Dacheng Tao, and Steven Hoi.
\newblock From images to textual prompts: Zero-shot visual question answering with frozen large language models.
\newblock In \emph{Proceedings of the IEEE/CVF Conference on Computer Vision and Pattern Recognition}, pages 10867--10877, 2023.

\bibitem[He et~al.(2016)He, Zhang, Ren, and Sun]{he2016deep}
Kaiming He, Xiangyu Zhang, Shaoqing Ren, and Jian Sun.
\newblock Deep residual learning for image recognition.
\newblock In \emph{Proceedings of the IEEE conference on computer vision and pattern recognition}, pages 770--778, 2016.

\bibitem[Hendrycks and Dietterich(2019)]{cifar100-C}
Dan Hendrycks and Thomas~G. Dietterich.
\newblock Benchmarking neural network robustness to common corruptions and perturbations.
\newblock In \emph{7th International Conference on Learning Representations, {ICLR} 2019, New Orleans, LA, USA, May 6-9, 2019}. OpenReview.net, 2019.
\newblock URL \url{https://openreview.net/forum?id=HJz6tiCqYm}.

\bibitem[Hendrycks and Gimpel(2017)]{hendrycks17baseline}
Dan Hendrycks and Kevin Gimpel.
\newblock A baseline for detecting misclassified and out-of-distribution examples in neural networks.
\newblock \emph{Proceedings of International Conference on Learning Representations}, 2017.

\bibitem[Hendrycks et~al.(2020)Hendrycks, Mu, Cubuk, Zoph, Gilmer, and Lakshminarayanan]{hendrycks2020augmix}
Dan Hendrycks, Norman Mu, Ekin~D. Cubuk, Barret Zoph, Justin Gilmer, and Balaji Lakshminarayanan.
\newblock {AugMix}: A simple data processing method to improve robustness and uncertainty.
\newblock \emph{Proceedings of the International Conference on Learning Representations (ICLR)}, 2020.

\bibitem[Jain et~al.(2023)Jain, Lawrence, Moitra, and Madry]{jain2023distilling}
Saachi Jain, Hannah Lawrence, Ankur Moitra, and Aleksander Madry.
\newblock Distilling model failures as directions in latent space.
\newblock In \emph{The Eleventh International Conference on Learning Representations}, 2023.
\newblock URL \url{https://openreview.net/forum?id=99RpBVpLiX}.

\bibitem[Jiang et~al.(2019)Jiang, Neyshabur, Mobahi, Krishnan, and Bengio]{jiang2019fantastic}
Yiding Jiang, Behnam Neyshabur, Hossein Mobahi, Dilip Krishnan, and Samy Bengio.
\newblock Fantastic generalization measures and where to find them.
\newblock \emph{arXiv preprint arXiv:1912.02178}, 2019.

\bibitem[Jiang et~al.(2022)Jiang, Nagarajan, Baek, and Kolter]{jiang2022assessing}
Yiding Jiang, Vaishnavh Nagarajan, Christina Baek, and J~Zico Kolter.
\newblock Assessing generalization of {SGD} via disagreement.
\newblock In \emph{International Conference on Learning Representations}, 2022.
\newblock URL \url{https://openreview.net/forum?id=WvOGCEAQhxl}.

\bibitem[Joshi et~al.(2022)Joshi, Pan, and He]{joshi2022all}
Nitish Joshi, Xiang Pan, and He~He.
\newblock Are all spurious features in natural language alike? an analysis through a causal lens.
\newblock \emph{arXiv preprint arXiv:2210.14011}, 2022.

\bibitem[Kirsch et~al.(2021)Kirsch, Mukhoti, van Amersfoort, Torr, and Gal]{kirsch2021pitfalls}
Andreas Kirsch, Jishnu Mukhoti, Joost van Amersfoort, Philip H.~S. Torr, and Yarin Gal.
\newblock On pitfalls in ood detection: Entropy considered harmful, 2021.
\newblock Uncertainty \& Robustness in Deep Learning Workshop, ICML.

\bibitem[Lee et~al.(2022)Lee, Chen, Tajwar, Kumar, Yao, Liang, and Finn]{lee2022surgical}
Yoonho Lee, Annie~S Chen, Fahim Tajwar, Ananya Kumar, Huaxiu Yao, Percy Liang, and Chelsea Finn.
\newblock Surgical fine-tuning improves adaptation to distribution shifts.
\newblock \emph{arXiv preprint arXiv:2210.11466}, 2022.

\bibitem[Li et~al.(2017)Li, Yang, Song, and Hospedales]{PACS}
Da~Li, Yongxin Yang, Yi-Zhe Song, and Timothy~M Hospedales.
\newblock Deeper, broader and artier domain generalization.
\newblock In \emph{Proceedings of the IEEE international conference on computer vision}, pages 5542--5550, 2017.

\bibitem[Li et~al.(2022)Li, Li, Xiong, and Hoi]{li2022blip}
Junnan Li, Dongxu Li, Caiming Xiong, and Steven Hoi.
\newblock Blip: Bootstrapping language-image pre-training for unified vision-language understanding and generation.
\newblock In \emph{International Conference on Machine Learning}, pages 12888--12900. PMLR, 2022.

\bibitem[Liu et~al.(2020)Liu, Wang, Owens, and Li]{liu2020energy}
Weitang Liu, Xiaoyun Wang, John Owens, and Yixuan Li.
\newblock Energy-based out-of-distribution detection.
\newblock \emph{Advances in neural information processing systems}, 33:\penalty0 21464--21475, 2020.

\bibitem[Liu et~al.(2015)Liu, Luo, Wang, and Tang]{liu2015faceattributes}
Ziwei Liu, Ping Luo, Xiaogang Wang, and Xiaoou Tang.
\newblock Deep learning face attributes in the wild.
\newblock In \emph{Proceedings of International Conference on Computer Vision (ICCV)}, December 2015.

\bibitem[Merullo et~al.(2022)Merullo, Castricato, Eickhoff, and Pavlick]{merullo2022linearly}
Jack Merullo, Louis Castricato, Carsten Eickhoff, and Ellie Pavlick.
\newblock Linearly mapping from image to text space.
\newblock \emph{arXiv preprint arXiv:2209.15162}, 2022.

\bibitem[Michels et~al.(2023)Michels, Adaloglou, Kaiser, and Kollmann]{michels2023contrastive}
Felix Michels, Nikolas Adaloglou, Tim Kaiser, and Markus Kollmann.
\newblock Contrastive language-image pretrained (clip) models are powerful out-of-distribution detectors.
\newblock \emph{arXiv preprint arXiv:2303.05828}, 2023.

\bibitem[Minderer et~al.(2021)Minderer, Djolonga, Romijnders, Hubis, Zhai, Houlsby, Tran, and Lucic]{minderer2021revisiting}
Matthias Minderer, Josip Djolonga, Rob Romijnders, Frances Hubis, Xiaohua Zhai, Neil Houlsby, Dustin Tran, and Mario Lucic.
\newblock Revisiting the calibration of modern neural networks.
\newblock \emph{Advances in Neural Information Processing Systems}, 34:\penalty0 15682--15694, 2021.

\bibitem[Ming et~al.(2022)Ming, Cai, Gu, Sun, Li, and Li]{ming2022delving}
Yifei Ming, Ziyang Cai, Jiuxiang Gu, Yiyou Sun, Wei Li, and Yixuan Li.
\newblock Delving into out-of-distribution detection with vision-language representations.
\newblock In Alice~H. Oh, Alekh Agarwal, Danielle Belgrave, and Kyunghyun Cho, editors, \emph{Advances in Neural Information Processing Systems}, 2022.
\newblock URL \url{https://openreview.net/forum?id=KnCS9390Va}.

\bibitem[Nakano et~al.(2021)Nakano, Hilton, Balaji, Wu, Ouyang, Kim, Hesse, Jain, Kosaraju, Saunders, et~al.]{nakano2021webgpt}
Reiichiro Nakano, Jacob Hilton, Suchir Balaji, Jeff Wu, Long Ouyang, Christina Kim, Christopher Hesse, Shantanu Jain, Vineet Kosaraju, William Saunders, et~al.
\newblock Webgpt: Browser-assisted question-answering with human feedback.
\newblock \emph{arXiv preprint arXiv:2112.09332}, 2021.

\bibitem[Narayanaswamy et~al.(2022)Narayanaswamy, Anirudh, Kim, Mubarka, Spanias, and Thiagarajan]{narayanaswamygengap}
Vivek Narayanaswamy, Rushil Anirudh, Irene Kim, Yamen Mubarka, Andreas Spanias, and Jayaraman~J. Thiagarajan.
\newblock Predicting the generalization gap in deep models using anchoring.
\newblock In \emph{ICASSP 2022 - 2022 IEEE International Conference on Acoustics, Speech and Signal Processing (ICASSP)}, pages 4393--4397, 2022.

\bibitem[Ng et~al.(2022)Ng, Cho, Hulkund, and Ghassemi]{ng2022predicting}
Nathan Ng, Kyunghyun Cho, Neha Hulkund, and Marzyeh Ghassemi.
\newblock Predicting out-of-domain generalization with local manifold smoothness.
\newblock \emph{arXiv preprint arXiv:2207.02093}, 2022.

\bibitem[Peng et~al.(2019)Peng, Bai, Xia, Huang, Saenko, and Wang]{domainnet}
Xingchao Peng, Qinxun Bai, Xide Xia, Zijun Huang, Kate Saenko, and Bo~Wang.
\newblock Moment matching for multi-source domain adaptation.
\newblock In \emph{Proceedings of the IEEE/CVF international conference on computer vision}, pages 1406--1415, 2019.

\bibitem[Pratt et~al.(2023)Pratt, Covert, Liu, and Farhadi]{pratt2023does}
Sarah Pratt, Ian Covert, Rosanne Liu, and Ali Farhadi.
\newblock What does a platypus look like? generating customized prompts for zero-shot image classification.
\newblock In \emph{Proceedings of the IEEE/CVF International Conference on Computer Vision}, pages 15691--15701, 2023.

\bibitem[Qu et~al.(2022)Qu, Li, Foo, Kuen, Gu, and Liu]{qu2022improving}
Haoxuan Qu, Yanchao Li, Lin~Geng Foo, Jason Kuen, Jiuxiang Gu, and Jun Liu.
\newblock Improving the reliability for confidence estimation.
\newblock In \emph{European Conference on Computer Vision}, pages 391--408. Springer, 2022.

\bibitem[Radford et~al.(2019)Radford, Wu, Child, Luan, Amodei, Sutskever, et~al.]{radford2019language}
Alec Radford, Jeffrey Wu, Rewon Child, David Luan, Dario Amodei, Ilya Sutskever, et~al.
\newblock Language models are unsupervised multitask learners.
\newblock \emph{OpenAI blog}, 1\penalty0 (8):\penalty0 9, 2019.

\bibitem[Radford et~al.(2021{\natexlab{a}})Radford, Kim, Hallacy, Ramesh, Goh, Agarwal, Sastry, Askell, Mishkin, Clark, et~al.]{clip}
Alec Radford, Jong~Wook Kim, Chris Hallacy, Aditya Ramesh, Gabriel Goh, Sandhini Agarwal, Girish Sastry, Amanda Askell, Pamela Mishkin, Jack Clark, et~al.
\newblock Learning transferable visual models from natural language supervision.
\newblock In \emph{International conference on machine learning}, pages 8748--8763. PMLR, 2021{\natexlab{a}}.

\bibitem[Radford et~al.(2021{\natexlab{b}})Radford, Kim, Hallacy, Ramesh, Goh, Agarwal, Sastry, Askell, Mishkin, Clark, et~al.]{radford2021learning}
Alec Radford, Jong~Wook Kim, Chris Hallacy, Aditya Ramesh, Gabriel Goh, Sandhini Agarwal, Girish Sastry, Amanda Askell, Pamela Mishkin, Jack Clark, et~al.
\newblock Learning transferable visual models from natural language supervision.
\newblock In \emph{International conference on machine learning}, pages 8748--8763. PMLR, 2021{\natexlab{b}}.

\bibitem[Russakovsky et~al.(2015)Russakovsky, Deng, Su, Krause, Satheesh, Ma, Huang, Karpathy, Khosla, Bernstein, Berg, and Fei-Fei]{imagenet}
Olga Russakovsky, Jia Deng, Hao Su, Jonathan Krause, Sanjeev Satheesh, Sean Ma, Zhiheng Huang, Andrej Karpathy, Aditya Khosla, Michael Bernstein, Alexander~C. Berg, and Li~Fei-Fei.
\newblock {ImageNet Large Scale Visual Recognition Challenge}.
\newblock \emph{International Journal of Computer Vision (IJCV)}, 115\penalty0 (3):\penalty0 211--252, 2015.
\newblock \doi{10.1007/s11263-015-0816-y}.

\bibitem[Schwenk et~al.(2022)Schwenk, Khandelwal, Clark, Marino, and Mottaghi]{schwenk2022okvqa}
Dustin Schwenk, Apoorv Khandelwal, Christopher Clark, Kenneth Marino, and Roozbeh Mottaghi.
\newblock A-okvqa: A benchmark for visual question answering using world knowledge.
\newblock In \emph{European Conference on Computer Vision}, pages 146--162. Springer, 2022.

\bibitem[Song et~al.(2022)Song, Dong, Zhang, Liu, and Wei]{song2022clip}
Haoyu Song, Li~Dong, Wei-Nan Zhang, Ting Liu, and Furu Wei.
\newblock Clip models are few-shot learners: Empirical studies on vqa and visual entailment.
\newblock \emph{arXiv preprint arXiv:2203.07190}, 2022.

\bibitem[Subramanyam et~al.(2023)Subramanyam, Jayram, Anirudh, and Thiagarajan]{subramanyam2023crepe}
Rakshith Subramanyam, TS~Jayram, Rushil Anirudh, and Jayaraman~J Thiagarajan.
\newblock Crepe: Learnable prompting with clip improves visual relationship prediction.
\newblock \emph{arXiv preprint arXiv:2307.04838}, 2023.

\bibitem[Touvron et~al.(2023)Touvron, Martin, Stone, Albert, Almahairi, Babaei, Bashlykov, Batra, Bhargava, Bhosale, et~al.]{touvron2023llama}
Hugo Touvron, Louis Martin, Kevin Stone, Peter Albert, Amjad Almahairi, Yasmine Babaei, Nikolay Bashlykov, Soumya Batra, Prajjwal Bhargava, Shruti Bhosale, et~al.
\newblock Llama 2: Open foundation and fine-tuned chat models.
\newblock \emph{arXiv preprint arXiv:2307.09288}, 2023.

\bibitem[Trivedi et~al.(2023)Trivedi, Koutra, and Thiagarajan]{trivedi2023closer}
Puja Trivedi, Danai Koutra, and Jayaraman~J Thiagarajan.
\newblock A closer look at scoring functions and generalization prediction.
\newblock In \emph{ICASSP 2023-2023 IEEE International Conference on Acoustics, Speech and Signal Processing (ICASSP)}, pages 1--5. IEEE, 2023.

\bibitem[Wang et~al.(2019)Wang, Ge, Lipton, and Xing]{imagenet-sketch}
Haohan Wang, Songwei Ge, Zachary Lipton, and Eric~P Xing.
\newblock Learning robust global representations by penalizing local predictive power.
\newblock In \emph{Advances in Neural Information Processing Systems}, pages 10506--10518, 2019.

\bibitem[Wang et~al.(2023)Wang, Li, Yao, and Li]{wang2023clipn}
Hualiang Wang, Yi~Li, Huifeng Yao, and Xiaomeng Li.
\newblock Clipn for zero-shot ood detection: Teaching clip to say no.
\newblock In \emph{Proceedings of the IEEE/CVF International Conference on Computer Vision}, pages 1802--1812, 2023.

\bibitem[Wei et~al.(2022)Wei, Bosma, Zhao, Guu, Yu, Lester, Du, Dai, and Le]{wei2022finetuned}
Jason Wei, Maarten Bosma, Vincent Zhao, Kelvin Guu, Adams~Wei Yu, Brian Lester, Nan Du, Andrew~M. Dai, and Quoc~V Le.
\newblock Finetuned language models are zero-shot learners.
\newblock In \emph{International Conference on Learning Representations}, 2022.
\newblock URL \url{https://openreview.net/forum?id=gEZrGCozdqR}.

\bibitem[Wei et~al.(2023)Wei, Hu, Xie, Liu, Zhang, Cao, Bao, Chen, and Guo]{wei2023improving}
Yixuan Wei, Han Hu, Zhenda Xie, Ze~Liu, Zheng Zhang, Yue Cao, Jianmin Bao, Dong Chen, and Baining Guo.
\newblock Improving clip fine-tuning performance.
\newblock In \emph{Proceedings of the IEEE/CVF International Conference on Computer Vision}, pages 5439--5449, 2023.

\bibitem[Wortsman et~al.(2022)Wortsman, Ilharco, Kim, Li, Kornblith, Roelofs, Lopes, Hajishirzi, Farhadi, Namkoong, et~al.]{wortsman2022robust}
Mitchell Wortsman, Gabriel Ilharco, Jong~Wook Kim, Mike Li, Simon Kornblith, Rebecca Roelofs, Raphael~Gontijo Lopes, Hannaneh Hajishirzi, Ali Farhadi, Hongseok Namkoong, et~al.
\newblock Robust fine-tuning of zero-shot models.
\newblock In \emph{Proceedings of the IEEE/CVF Conference on Computer Vision and Pattern Recognition}, pages 7959--7971, 2022.

\bibitem[Yang et~al.(2023)Yang, Zhang, Katabi, and Ghassemi]{yang2023change}
Yuzhe Yang, Haoran Zhang, Dina Katabi, and Marzyeh Ghassemi.
\newblock Change is hard: A closer look at subpopulation shift.
\newblock In \emph{International Conference on Machine Learning}, 2023.

\bibitem[Yu et~al.(2024)Yu, Cho, Yadav, and Bansal]{yu2024self}
Shoubin Yu, Jaemin Cho, Prateek Yadav, and Mohit Bansal.
\newblock Self-chained image-language model for video localization and question answering.
\newblock \emph{Advances in Neural Information Processing Systems}, 36, 2024.

\bibitem[Yun et~al.(2019)Yun, Han, Oh, Chun, Choe, and Yoo]{yun2019cutmix}
Sangdoo Yun, Dongyoon Han, Seong~Joon Oh, Sanghyuk Chun, Junsuk Choe, and Youngjoon Yoo.
\newblock Cutmix: Regularization strategy to train strong classifiers with localizable features.
\newblock In \emph{International Conference on Computer Vision (ICCV)}, 2019.

\bibitem[Zhu et~al.(2022)Zhu, Cheng, Zhang, and Liu]{zhu2022rethinking}
Fei Zhu, Zhen Cheng, Xu-Yao Zhang, and Cheng-Lin Liu.
\newblock Rethinking confidence calibration for failure prediction.
\newblock In \emph{European Conference on Computer Vision}, pages 518--536. Springer, 2022.

\end{thebibliography}
\newpage

\appendix
\section{Additional Analysis of \method}

\subsection{Impact of PIM's predictive performance}

We note that, the quality of attributes has a direct impact on PIM.
\begin{table}[h!]
\centering
\caption{Regardless of the predictive performance w.r.t. the task model, PIM is very useful for failure detection.}
\renewcommand{\arraystretch}{1.5}
\resizebox{\columnwidth}{!}{
\begin{tabular}{cc|cc|cc|}
\cline{3-6}
                                                              & \textbf{}       & \multicolumn{2}{c|}{\textbf{Accuracy (\%)}}              & \multicolumn{2}{c|}{\textbf{Failure Recall}}           \\ \hline
\multicolumn{1}{|c|}{{Source}} & {Target} & \multicolumn{1}{c|}{{Baseline}} & {PIM}   & \multicolumn{1}{c|}{{Baseline}} & {PIM}  \\ \hline
\multicolumn{1}{|c|}{Waterbirds}                               &   Waterbirds  & \multicolumn{1}{c|}{{75.06}}                            & {\color[HTML]{3531FF} \textbf{82.11}  }      & \multicolumn{1}{c|}{0.49}                                      & {\color[HTML]{3531FF} \textbf{0.61}} \\ \hline
\multicolumn{1}{|c|}{CIFAR-100}                               & CIFAR-100C      & \multicolumn{1}{c|}{{\color[HTML]{3531FF} \textbf{31.84}} }                           & 31.25          & \multicolumn{1}{c|}{0.81}                                      & {\color[HTML]{3531FF} \textbf{0.85}} \\ \hline
\multicolumn{1}{|c|}{DomainNet-R}                             & DomainNet-C     & \multicolumn{1}{c|}{{\color[HTML]{3531FF} \textbf{33.14}}}                            & 30.12          & \multicolumn{1}{c|}{0.68}                                      & {\color[HTML]{3531FF} \textbf{0.84}} \\ \hline
\multicolumn{1}{|c|}{DomainNet-R}                             & DomainNet-S     & \multicolumn{1}{c|}{{\color[HTML]{3531FF} \textbf{20.02}}}                            & {\color[HTML]{3531FF} \textbf{20.11}} & \multicolumn{1}{c|}{0.70}                                      & {\color[HTML]{3531FF} \textbf{0.85}} \\ \hline
\multicolumn{1}{|c|}{DomainNet-R}                             & DomainNet-P     & \multicolumn{1}{c|}{{31.25}}                            & {\color[HTML]{3531FF} \textbf{31.72}} & \multicolumn{1}{c|}{0.71}                                      & {\color[HTML]{3531FF} \textbf{0.77}} \\ \hline
\multicolumn{1}{|c|}{DomainNet-R}                             & DomainNet-I    & \multicolumn{1}{c|}{{11.16}}                            & {\color[HTML]{3531FF} \textbf{12.63} }& \multicolumn{1}{c|}{0.70}                                      &{\color[HTML]{3531FF}  \textbf{0.84}} \\ \hline
% \multicolumn{1}{|c|}{ImageNet}                                & Imagenet-R      & \multicolumn{1}{c|}{}                                          &                & \multicolumn{1}{c|}{}                                          &               \\ \hline
\multicolumn{1}{|c|}{ImageNet}                                & ImageNet-Sketch      & \multicolumn{1}{c|}{{\color[HTML]{3531FF} \textbf{23.57}}}                                          &  22.49              & \multicolumn{1}{c|}{0.90}                                          &  {\color[HTML]{3531FF} \textbf{0.95}  }           \\ \hline
\end{tabular}
}

\label{tab:pim_acc}

\end{table}That said, even with slightly lower performance, the core-nuisance attribute disambiguation, which is the most critical for failure detection, is not compromised. 
To demonstrate this, we show in Table~\ref{tab:pim_acc} the failure recall performance on Waterbirds, CIFAR-100, ImageNet, and DomainNet (Real to Clipart, Sketch, Painting, Infograph domains) datasets. Regardless of its predictive performance, the failure recall of \method is consistently higher than the best performing baseline.

\subsection{Replacing PIM with CLIP classifiers}
Through PIM, we create a variant of the task model that disambiguates core attributes (identified using LLM) from nuisance attributes. This means that if the task model and PIM do not agree on a prediction, it is likely a failure. However, with a zero-shot CLIP classifier, the visual features and their correlations to the core attributes can be drastically different from the task model. This renders the model disagreement based failure detection highly ineffective. To show this, we tried two versions of the CLIP classifier: one using text prompts for each class label (CLIP-cls) and another using core attributes (CLIP-att) like \method. From Table~\ref{tab:CLIP_PIM}, we see that PIM performs much better than both versions.

\begin{table}[h]
    \centering
     \caption{ Replacing PIM with CLIP classifiers}
    \resizebox{0.65\columnwidth}{!}{
    \begin{tabular}{c|c|c|c|}
    \cline{2-4}
        & CLIP-cls & CLIP-att & PIM \\
        \cline{1-4}
        \multicolumn{1}{|c|}{\method~MCC} & -0.01 & 0.02 & {\color[HTML]{3531FF} \textbf{0.46}}\\
        \hline
    \end{tabular}
    }
    \label{tab:CLIP_PIM}
\end{table}

\subsection{Ablating synthetic augmentations for PIM training }
We conducted an ablation study on CIFAR100 to demonstrate the value of augmentations used during PIM training. 
\begin{table}[h!]
    \centering
    \caption{\textbf{Impact of Augmentations}: We compare the failure detection performance in the presence and absence of augmentations for PIM training.}
    \renewcommand{\arraystretch}{1}
    \resizebox{0.8\columnwidth}{!}{
    \begin{tabular}{c|c|c|}
    \cline{2-3}
        & No Aug. &  Augmix (0.2) + Cutmix (0.2) \\
        \cline{1-3}
        \multicolumn{1}{|c|}{\method~MCC} & 0.27 & {\color[HTML]{3531FF}\textbf{0.53}}\\
        \hline
    \end{tabular}
    }
    \label{tab:aug}
\end{table}
As Table~\ref{tab:aug} shows, there is a significant drop in failure detection performance without augmentations.
\section{Algorithm Listing for PIM}
\begin{algorithm*}[h]
\caption{Training Procedure for Prior Induced Model $\phi$}
\begin{algorithmic}[1]
\Require Training data $\mathcal{D} = \{(\mathrm{x}_i, \mathrm{y}_i)\}_{i=1}^M$, attribute set $a_k$ for each class $c \in \mathcal{Y}$ extracted from LLM, VLM (CLIP) text encoder $T(.)$, classifier $\mathcal{F}$, cross-entropy loss $\mathcal{L}(.)$ parameters $\phi$ initialized with ImageNetv1 weights.
\Ensure Optimized parameters $\phi$.
\For{each epoch}
    \For{each batch $\{(\mathrm{x}_i, \mathrm{y}_i)\}$ in $\mathcal{D}$} 
        \State Apply augmentation (Cutmix or Augmix) across batch with probability $p$
        \State Compute features $\mathbf{h}_l$ for layer $l$ from $\mathcal{F}$.
        \State Use PIM to map $\mathbf{h}_l$ on to the VLM latent space to obtain $\phi(\mathbf{h}_l)$ 
        \State Initialize sample-level loss weights to uniform
        \For{each class $c$}
            \State Compute cosine similarity between $\phi(\mathbf{h}_l)$ and the CLIP text embeddings $T(.)$ of attributes from class $c$.
            \State Aggregate similarities to derive class-level logits $\Tilde{\mathrm{y}}$.
        \EndFor
        \State Compute sample-level loss weights based on the discrepancy in predictions between $\mathcal{F}$ and $\phi$.

        \State Update $\phi$ using the objective - $\min_{\phi}\mathcal{L}(\mathrm{y},\Tilde{\mathrm{y}})$
    \EndFor
\EndFor
\State \textbf{return} Optimized $\phi$.
\end{algorithmic}
\end{algorithm*}

\section{Training Details}

% Our codes can be accessed through 
% \url{https://anonymous.4open.science/r/PRIME-96EC}

\subsection{Classifier Training}
Table \ref{tab:training_params} provides the hyper-parameter and optimization settings for every dataset employed for training the classifier $\mathcal{F}$. 
We use a multi-step LR decay scheduler, which reduces the learning rate by a factor of $0.2$. 
% \noindent \textbf{CIFAR100:} Training spans 200 epochs, initial learning rate 0.1, multi-step decay at epochs 60, 120, 160, with tenfold reductions. Optimizer: SGD, momentum 0.2, weight decay 5e-4.

% \noindent \textbf{Waterbirds:} Training spans 100 epochs, initial learning rate 0.001, multi-step decay at epochs 30, 60, with tenfold reductions. Optimizer: SGD, momentum 0.9.

% \noindent \textbf{CelebA:} Training spans 20 epochs, learning rate 0.1. Optimizer: SGD, momentum 0.9.

% \noindent \textbf{Cats \& Dogs:} Training spans 100 epochs, initial learning rate 0.01, multi-step decay at epochs 30, 60, with tenfold reductions. Optimizer: SGD, momentum 0.9.

% \noindent \textbf{PACS:} Training spans 200 epochs, initial learning rate 0.01, multi-step decay at epochs 30, 60, with tenfold reductions. Optimizer: SGD, momentum 0.9.

\begin{table}[ht!]
\centering
\caption{Hyper-parameter and optimization settings for training classifier $\mathcal{F}$ for different datasets.}
\renewcommand{\arraystretch}{1.5} % Adjust the row height
\begin{tabular}{|l|c|c|c|c|c|}
\hline
Dataset      & Epochs & Initial Learning Rate & LR Decay epochs        & Momentum & Optimizer \\ \hline
CIFAR100     & 200    & 0.1                   & 60, 120, 160            & 0.2      & SGD       \\ 
Waterbirds   & 100    & 0.001                 & 30, 60                  & 0.9      & SGD       \\ 
CelebA       & 20     & 0.1                   & -                       & 0.9      & SGD       \\ 
Cats \& Dogs & 100    & 0.01                  & 30, 60                  & 0.9      & SGD       \\ 
PACS         & 200     &  0.01                & 60, 120 , 160        & 0.9      & SGD       \\ 
Domainnet & 30 &0.001 & -& 0.9 & Adam\\\hline

\end{tabular}

\label{tab:training_params}
\end{table}

\subsection{PIM (Prior Induced Model) Training Details}

We adopt the following protocol to train PIM for all datasets. We train PIM for 200 epochs, starting with an initial learning rate of 0.1, and implement multi-step decay at epochs 60, 120, and 160, where we reduce the learning rate by factor of 0.1 . We utilize the AdamW optimizer for optimization. Additionally, we apply both CutMix and AugMix transformations to the entire batch with probabilities of 0.2 each. Moreover, we carefully weight our loss function during training. The loss weights are increased by a factor of 2.0 for samples where the classifier $\mathcal{F}$ succeeds but PIM fails, and by 1.5 for cases where both the classifier and PIM fail.

% We adopt the follwing protocol to train PIM for all datasets. Training spans 200 epochs, initial learning rate of 0.1, and multi-step decay at epochs 60, 120, and 160, with tenfold reductions. Optimizer: AdamW. CutMix is applied to the entire batch with a probability of 0.2, and similarly, AugMix transformations are applied to the entire batch with a probability of 0.2. The loss is weighted 2.0 times for samples where the task model succeeds but PIM fails, and 1.5 times where both the classifier and PIM fail.

\section{Prompts Used to Query LLM (GPT3) for Attribute Generation}

\begin{itemize}
    \item \textbf{Waterbirds:} \textit{"List 100 distinct two-word phrases that uniquely describe the visual characteristics (like type of feet, beak, wings, plumage, feathers, feather texture, body shape, body type etc) of \{class\_name\}. Make sure the phrases are not long descriptions."}
    \item \textbf{CIFAR100:} \textit{"List 50 distinct two-word phrases that uniquely describe the visual characteristics (like shape, color, texture) of \{class\_name\}. Make sure the phrases are not long descriptions."}
    \item \textbf{PACS:} \textit{"List 30 distinct two-word phrases that uniquely describe the visual characteristics of \{class\_name\}. Do not describe their colors. Make sure the phrases are not long descriptions."}
    % \item \textbf{ImageNet:} \textit{"List 3 distinct two-word phrases that uniquely describe the visual characteristics of \{class\_name\}. Make sure the phrases are not long descriptions."}
    \item \textbf{CelebA:} \textit{"List 25 distinct two-word phrases that uniquely describe the visual characteristics of \{class\_name\} hair person. Make sure the phrases are not long descriptions."}
    \item \textbf{Cats and Dogs:} \textit{"List 50 distinct two-word phrases that uniquely describe the visual characteristics of \{class\_name\}. Make sure the phrases are not long descriptions."}
    \item \textbf{DomainNet:} \textit{"List 100 distinct two-word phrases that uniquely describe the visual characteristics of \{class\_name\}. Make sure the phrases are not long descriptions."}
    \item \textbf{ImageNet:}\textit{"List 100 distinct two-word phrases that uniquely describe the visual characteristics of \{class\_name\}. Make sure the phrases are not long descriptions."}
\end{itemize}

\section{Additional Results}

\noindent \textbf{Experiment with ViT-B-16}:  We extend our study to incorporate the ViT architecture, specifically using the ViT-B-16 model, for the Waterbirds datasets. We provide these results in \Cref{tab:vit}. For ViT-B-16, we explore two PIM variations: one using features from the first layer and another from the ninth layer of the ViT-B-16 classifier model. From \Cref{tab:vit} it is evident that \method~continues to outperform as a more reliable failure estimator, indicating its adaptability with various classifier architectures. Moreover, obtaining features from the initial layers of the classifier for constructing the PIM (Prior Induced Model) proves to be more effective than sourcing them from the deeper layers, aligning with our previous findings.

\begin{table*}[ht]
  \centering
  \caption{Performance Comparison for ViT-B-16 architecture on the Waterbirds dataset}
  \label{tab:vit}
 \resizebox{0.75\linewidth}{!}{
    \begin{tabular}{@{}l|cccccc@{}}
    \toprule
    \multicolumn{1}{c}{\textbf{Dataset}} & \textbf{Method} & \textbf{FR} & \textbf{SR} & \textbf{MCC} \\
    \midrule
\multirow{5}{*}{Waterbirds} 
          &  \msp & 0.1587 & 0.9048 & 0.0954\\
          &  Energy & 0.4924 & 0.6732 & 0.1656\\
          &  \ent & 0.3076 & 0.8301 & 0.1613\\
    \cmidrule(lr){2-7}
          & \multicolumn{1}{l}{~\method~layer1} & & & & & \\
          & \phantom{xx}+ mean     &   0.5592 & 0.7743 & 0.3416\\
          & \phantom{x}+ max & 0.6056 & 0.8091 & \secondbest{0.4235}\\

          & \multicolumn{1}{l}{~\method~layer9} & & & & & \\
          & \phantom{xx}+ mean     &   0.4818 & 0.6789 & 0.1613\\
          & \phantom{x}+ max & 0.5380 & 0.7542 & 0.2970\\

    % \midrule
    % \multirow{5}{*}{Cats \& Dogs} 
    %       &  \msp & 0.3062 & 0.8910 & 0.2317\\
    %       &  Energy & 0.4236 & 0.8329 & 0.2606\\
    %       &  \ent & 0.3817 & 0.8468 & 0.2402\\
    % \cmidrule(lr){2-7}
    %       & \multicolumn{1}{l}{~\method~layer1} & & & & & \\
    %       & \phantom{xx}+ mean &   0. & 0. & 0.\\
    %       & \phantom{x}+ max & 0. & 0. & \secondbest{0.}\\

    %       & \multicolumn{1}{l}{~\method~layer9} & & & & & \\
    %       & \phantom{xx}+ mean &   0. & 0. & 0.\\
    %       & \phantom{x}+ max & 0.2256 & 0.8455 & 0.0805\\

    \end{tabular}%
    }
\end{table*}%

\noindent \textbf{Detailed Results with PACS}: Expanding on the results provided in the main paper, we provide the failure detection performance metrics under the settings where classifier $\mathcal{F}$ and PIM $\phi$ are trained on different domains. For all experiments, we used early layer features of the classifier. For each of these experiments, the failure estimation threshold is established based on the validation set from the respective training domain. The additional results are tabulated in ~\Cref{tab:pacs_art_painting} to ~\Cref{tab:pacs_sketch}.

% Building upon our initial results for the PACS dataset from the main paper,  

\begin{table*}[ht]
  \centering
  \caption{Performance Comparison on PACS dataset, where the classifier and the PIM are trained and calibrated on the \textit{Art Painting} domain}
  \label{tab:pacs_art_painting}
 \resizebox{0.75\linewidth}{!}{
    \begin{tabular}{@{}l|cccccc@{}}
    \toprule
    \multicolumn{1}{c}{\textbf{Eval. Domain}} & \textbf{Method} & \textbf{FR} & \textbf{SR} & \textbf{MCC} \\
    \midrule
\multirow{5}{*}{Art Painting} 
          &  \msp & 0.7345 & 0.7799 & 0.4564\\
          &  Energy & 0.6381 & 0.7698 & 0.3659\\
          &  \ent & 0.6959 & 0.7837 & 0.4294\\
    \cmidrule(lr){2-7}
          & \multicolumn{1}{l}{~\method} & & & & & \\
          & \phantom{xx}+ mean     &   0.7516 & 0.9822 & 0.7928\\
          & \phantom{x}+ max & 0.8458 & 0.9784 & \secondbest{0.8498}\\
    \midrule

    \multirow{5}{*}{Cartoon } 
          &  \msp & 0.5636 & 0.6675 & 0.2204\\
          &  Energy & 0.5033 & 0.7188 & 0.2141\\
          &  \ent & 0.5799  & 0.6687 & 0.2371\\
    \cmidrule(lr){2-7}
          & \multicolumn{1}{l}{~\method} & & & & & \\
          & \phantom{xx}+ mean  & 0.8394 & 0.6430 & 0.4895\\
          & \phantom{x}+ max  & 0.8945 & 0.6027 & \secondbest{0.5284}\\

    \midrule

    \multirow{5}{*}{Photo } 
          &  \msp & 0.5660 & 0.7857 & 0.3617\\
          &  Energy & 0.5406 & 0.8265 & 0.3851\\
          &  \ent & 0.5305  & 0.8254 & 0.3743\\
    \cmidrule(lr){2-7}
          & \multicolumn{1}{l}{~\method} & & & & & \\
          & \phantom{xx}+ mean  & 0.6942 & 0.8594 & 0.5637\\
          & \phantom{x}+ max  & 0.7754 & 0.8424 & \secondbest{0.6200}\\

    \midrule
    \multirow{5}{*}{Sketch } 
          &  \msp & 0.6412 & 0.6088 & 0.2445\\
          &  Energy & 0.3252 & 0.8238 & 0.1639\\
          &  \ent & 0.6001  & 0.6252 & 0.2196\\
    \cmidrule(lr){2-7}
          & \multicolumn{1}{l}{~\method} & & & & & \\
          & \phantom{xx}+ mean  & 0.8642 & 0.4977 & 0.3944\\
          & \phantom{x}+ max  & 0.9066 & 0.4576 & \secondbest{0.4187}\\

    \end{tabular}%
    }
\end{table*}%

\begin{table*}[ht]
  \centering
  \caption{Performance Comparison on PACS dataset, where the classifier and the PIM are trained and calibrated on \textit{Cartoon} domain}
  \label{tab:pacs_cartoon}
 \resizebox{0.75\linewidth}{!}{
    \begin{tabular}{@{}l|cccccc@{}}
    \toprule
    \multicolumn{1}{c}{\textbf{Eval. Domain}} & \textbf{Method} & \textbf{FR} & \textbf{SR} & \textbf{MCC} \\

    \midrule
    \multirow{5}{*}{Art Painting} 
          &  \msp & 0.4988 & 0.6999 & 0.1938\\
          &  Energy & 0.5467 & 0.6335 & 0.1739\\
          &  \ent & 0.4772 & 0.7118 & 0.1855\\
    \cmidrule(lr){2-7}
          & \multicolumn{1}{l}{~\method} & & & & & \\
          & \phantom{xx}+ mean     &   0.6602 & 0.7556 & \secondbest{0.4011}\\
          & \phantom{x}+ max & 0.6270 & 0.7849 & 0.3977\\
    \midrule

    \multirow{5}{*}{Cartoon } 
          &  \msp & 0.6280 & 0.9206 & 0.4343 \\
          &  Energy & 0.5427 & 0.9211 & 0.3761 \\
          &  \ent & 0.5061 & 0.9349 & 0.3818  \\
    \cmidrule(lr){2-7}
          & \multicolumn{1}{l}{~\method} & & & & & \\
          & \phantom{xx}+ mean  & 0.6341 & 0.9950 & \secondbest{0.7430}\\
          & \phantom{x}+ max  & 0.5854 & 0.9950 & 0.7092\\

    \midrule

    \multirow{5}{*}{Photo } 
          &  \msp & 0.4561 & 0.7660 & 0.2281\\
          &  Energy & 0.4819 & 0.7418 & 0.2266\\
          &  \ent & 0.4355 & 0.7974 & 0.2431\\
    \cmidrule(lr){2-7}
          & \multicolumn{1}{l}{~\method} & & & & & \\
          & \phantom{xx}+ mean  & 0.6656 & 0.8916 & \secondbest{0.5552}\\
          & \phantom{x}+ max  & 0.6316 & 0.9016 & 0.5354\\

    \midrule
    \multirow{5}{*}{Sketch } 
          &  \msp & 0.6033 & 0.6570 & 0.2497\\
          &  Energy & 0.5668 & 0.7372 & 0.2926\\
          &  \ent & 0.5470 & 0.7202 & 0.2575\\
    \cmidrule(lr){2-7}
          & \multicolumn{1}{l}{~\method} & & & & & \\
          & \phantom{xx}+ mean  & 0.7604 & 0.8871 & \secondbest{0.6220}\\
          & \phantom{x}+ max  & 0.7132 & 0.9006 & 0.5887\\

    \end{tabular}%
    }
\end{table*}%

\begin{table*}[ht]
  \centering
  \caption{Performance Comparison on PACS dataset, where the classifier and the PIM are trained and calibrated on \textit{Photo} domain}
  \label{tab:pacs_photo}
 \resizebox{0.75\linewidth}{!}{
    \begin{tabular}{@{}l|cccccc@{}}
    \toprule
    \multicolumn{1}{c}{\textbf{Eval. Domain}} & \textbf{Method} & \textbf{FR} & \textbf{SR} & \textbf{MCC} \\

    \midrule
    \multirow{5}{*}{Art Painting} 
          &  \msp & 0.5364 & 0.5983 & 0.1220\\
          &  Energy & 0.6269 & 0.5254 & 0.1399\\
          &  \ent & 0.5658 & 0.5847 & 0.1365\\
    \cmidrule(lr){2-7}
          & \multicolumn{1}{l}{~\method} & & & & & \\
          & \phantom{xx}+ mean     &   0.6272 & 0.6955 & \secondbest{0.2913}\\
          & \phantom{x}+ max & 0.5653 & 0.7266 & 0.2630\\
    \midrule

    \multirow{5}{*}{Cartoon } 
          &  \msp & 0.43532 & 0.56802 & 0.00258 \\
          &  Energy & 0.59117 & 0.51313 & 0.08078 \\
          &  \ent & 0.44831 & 0.55131 & -0.00029  \\
    \cmidrule(lr){2-7}
          & \multicolumn{1}{l}{~\method} & & & & & \\
          & \phantom{xx}+ mean  & 0.47292 & 0.66274 & 0.10499\\
          & \phantom{x}+ max  & 0.42448 & 0.75236 & \secondbest{0.13940}\\

    \midrule

    \multirow{5}{*}{Photo } 
            & \msp & 0.5278 & 0.9835 & 0.4537 \\
            & Energy & 0.5000 & 0.9633 & 0.3189 \\
            & \ent & 0.5556 & 0.9859 & 0.4965 \\
    \cmidrule(lr){2-7}
          & \multicolumn{1}{l}{~\method} & & & & & \\
          & \phantom{xx}+ mean  & 0.7143 & 0.9939 & \secondbest{0.7082}\\
          & \phantom{x}+ max  & 0.6571 & 0.9927 & 0.6498\\

    \midrule
    \multirow{5}{*}{Sketch } 
        & \msp & 0.2226 & 0.8689 & 0.0886 \\
        & Energy & 0.3440 & 0.9324 & 0.2377 \\
        & \ent & 0.2176 & 0.8919 & 0.1077 \\
    \cmidrule(lr){2-7}
          & \multicolumn{1}{l}{~\method} & & & & & \\
          & \phantom{xx}+ mean  & 0.4229 & 0.9424 & \secondbest{0.2996} \\
          & \phantom{x}+ max  & 0.4103 & 0.9263 & 0.2774 \\

    \end{tabular}%
    }
\end{table*}%

\begin{table*}[ht]
  \centering
  \caption{Performance Comparison on PACS dataset, where the classifier and the PIM are trained and calibrated on \textit{Sketch} domain}
  \label{tab:pacs_sketch}
 \resizebox{0.75\linewidth}{!}{
    \begin{tabular}{@{}l|cccccc@{}}
    \toprule
    \multicolumn{1}{c}{\textbf{Eval. Domain}} & \textbf{Method} & \textbf{FR} & \textbf{SR} & \textbf{MCC} \\

    \midrule
    \multirow{5}{*}{Art Painting} 
        & \msp & 0.3836 & 0.6026 & -0.0112 \\
        & Energy & 0.3317 & 0.6462 & -0.0184 \\
        & \ent & 0.4331 & 0.5513 & -0.0124 \\
    \cmidrule(lr){2-7}
          & \multicolumn{1}{l}{~\method} & & & & & \\
          & \phantom{xx}+ mean     &   0.9156 & 0.1769 & 0.1200 \\
          & \phantom{x}+ max & 0.9710 & 0.1179 & \secondbest{0.1670} \\
    \midrule

    \multirow{5}{*}{Cartoon } 
        & \msp & 0.4536 & 0.6892 & 0.1326 \\
        & Energy & 0.5270 & 0.6129 & 0.1279 \\
        & \ent & 0.5215 & 0.6633 & 0.1692 \\
    \cmidrule(lr){2-7}
          & \multicolumn{1}{l}{~\method} & & & & & \\
          & \phantom{xx}+ mean  & 0.8830 & 0.5640 & \secondbest{0.4717} \\
          & \phantom{x}+ max  & 0.8563 & 0.5338 & 0.4065 \\

    \midrule

    \multirow{5}{*}{Photo } 
        & \msp & 0.3107 & 0.6667 & -0.0179 \\
        & Energy & 0.2750 & 0.7074 & -0.0145 \\
        & \ent & 0.3479 & 0.6481 & -0.0031 \\
    \cmidrule(lr){2-7}
          & \multicolumn{1}{l}{~\method} & & & & & \\
          & \phantom{xx}+ mean  & 0.9679 & 0.1333 & 0.1734 \\
          & \phantom{x}+ max  & 0.9850 & 0.1148 & \secondbest{0.2116} \\
          
    \midrule
    \multirow{5}{*}{Sketch } 
            & \msp & 0.6822 & 0.9532 & 0.4221 \\
            & Energy & 0.3458 & 0.9314 & 0.1702 \\
            & \ent & 0.6449 & 0.9464 & 0.3778 \\
    \cmidrule(lr){2-7}
          & \multicolumn{1}{l}{~\method} & & & & & \\
          & \phantom{xx}+ mean  & 0.4673 & 0.9950 & \secondbest{0.5729} \\
          & \phantom{x}+ max  & 0.4299 & 0.9639 & 0.3034 \\

    \end{tabular}%
    }
\end{table*}%

\end{document}